\documentclass[runningheads]{llncs}
\usepackage[T1]{fontenc}
\usepackage{graphicx}
\usepackage{cite}
\usepackage{amsmath,amssymb,amsfonts}
\usepackage{algpseudocode}
\usepackage{algorithm2e}
\usepackage{graphicx}
\usepackage{textcomp}
\usepackage{xcolor}
\usepackage{epsfig}
\usepackage{tabularx}
\usepackage{booktabs}
\usepackage{verbatim}
\usepackage{xcolor}
\usepackage{amstext}
\usepackage{upgreek}
\usepackage{multirow}
\usepackage{bbm}
\usepackage{derivative}

\usepackage{wrapfig}

\usepackage{enumitem}
	
\usepackage{color, colortbl}

\usepackage{lipsum}  

\newcommand{\sskip}{\mskip1mu}

\definecolor{Gray}{gray}{0.9}

\newcommand{\mr}[1]{\text{#1}}

\newcommand{\mypar}[1]{\vspace{0.25em}\noindent\textbf{#1}~}
\newcommand{\change}[1]{#1}
\begin{document}
%
\title{Mixup-Privacy: A simple yet effective approach for privacy-preserving segmentation \thanks{Supported by Natural Sciences and Engineering Research Council of Canada (NSERC) and Reseau de BioImagrie du Quebec (RBIQ)}}
\titlerunning{Mixup-Privacy}
%
\author{Bach Ngoc Kim\inst{1}\orcidID{0000-0001-6188-9358} \and
Jose Dolz\inst{1}\orcidID{0000-0002-2436-7750} \and
Pierre-Marc Jodoin\inst{2}\orcidID{0000-0002-6038-5753} \and
Christian Desrosiers\inst{1}\orcidID{0000-0002-9162-9650}}
%
\authorrunning{B. N. Kim et al.}
%
\institute{École de Technologie Supérieure, Montreal QC H3C1K3, Canada \and
Université de Sherbrooke, Sherbrooke QC J1K2R1, Canada}
%
\maketitle              
\begin{abstract}
Privacy protection in medical data is a legitimate obstacle for centralized machine learning applications. Here, we propose a client-server image segmentation system which allows for the analysis of multi-centric medical images while preserving patient privacy. In this approach, the client protects the to-be-segmented patient image by mixing it to a reference image. As shown in our work, it is \change{challenging} to separate the image mixture to exact original content, thus making the data unworkable and unrecognizable for an unauthorized person. This proxy image is sent to a server for processing. The server then returns the mixture of segmentation maps, which the client can revert to a correct target segmentation. Our system has two components: 1) a segmentation network on the server side which processes the image mixture, and 2) a segmentation unmixing network which recovers the correct segmentation map from the segmentation mixture. Furthermore, the whole system is trained end-to-end.
The proposed method is validated on the task of MRI brain segmentation using images from two different datasets. Results show that the segmentation accuracy of our method is comparable to a system trained on raw images, and outperforms other privacy-preserving methods with little computational overhead. \vspace{-0.2cm}
\vspace{-0.1cm}
\vspace{-0.3cm}
\end{abstract}
\section{Introduction}
\vspace{-0.3cm}
Neural networks are the {\em de facto} solution to numerous medical analysis tasks, from disease recognition, to anomaly detection, segmentation, tumor resurgence prediction, and many more~\cite{dolz20183d,litjens2017survey}. Despite their success, the widespread clinical deployment of neural nets has been hindered by legitimate privacy restrictions, which limit the amount of data the scientific community can pool together.

Researchers have explored a breadth of solutions to tap into massive \change{amounts} of data while complying with privacy restrictions. One such solution is federated learning (FL)~\cite{DBLP:journals/corr/KonecnyMRR16,yang2019federated}, for which training is done across a network of computers each \change{holding} its local data. While FL has been shown \change{to be} effective, it nonetheless suffers from \change{some limitations} when it comes to medical data. First, from a cybersecurity standpoint, communicating with computers located in a highly-secured environment such as a hospital, while complying with FDA/MarkCE cybersecurity regulation, is no easy feast. Second, having computers communicate with their local PACS server is also tricky. And third, since FL is a decentralized {\em training} solution, it requires a decentralized set of computers to process images at test time, making it ill-suited for software as a service (SAAS) cloud services. Another solution is to train a centralized network with homomorphic data encryption~\cite{Hardy2017}.  While this ensures a rigorous data protection, as detailed in Section~\ref{sec:rw}, the tremendous computational complexity of homomorphic networks prohibits their use in practice.

Recent studies have investigated centralized cloud-based solutions where data is encoded by a neural network prior \change{being sent} to the server~\cite{kim2020privacynet}.  While the encoded data is unworkable for unauthorized parties, it nonetheless can be processed by a network that was trained to deal with such encoded data. In some methods, such as Privacy-Net~\cite{kim2020privacynet}, the data sent back to the client (e.g., predicted segmentation maps) is not encoded and may contain some private information about the patient (e.g., the patient's identity or condition). To ensure that the returned data is also unworkable for non-authorized users, Kim et al.\cite{bach2021nonlinear} proposed an encoding method based on reversible image warping, where the warping function is only known by the client.

In this paper, we propose a novel client-server cloud system that can effectively segment medical images while protecting subjects' data privacy. Our segmentation method, which relies on the hardness of blind source separation (BSS) as root problem~\cite{article_BSS,720250,NOURI202275,DAVIES20071819}, leverages a simple yet powerful technique based on mixup~\cite{guo2019mixup}. In the proposed approach, the client protects the to-be-segmented patient image by mixing it to a reference image only known to this client. This reference image can be thought as a private key needed to encode and decode the image and its segmentation map. The image mixture renders the data unworkable and unrecognizable for a non-authorized person, since recovering the original images requires to solve an intractable BSS problem. This proxy image is sent to a server for a processing task, which corresponds to semantic segmentation in this work. Instead of sending back the non-encoded segmentation map, as in~\cite{kim2020privacynet}, the server returns to the client a mixture of the target and reference segmentation maps. Finally, because the client knows the segmentation map for the reference image, as well as the mixing coefficients, it can easily recover the segmentation for the target. 

Our work makes four contributions to privacy-preserving segmentation:
\begin{enumerate}[topsep=0pt,itemsep=1.5pt]
\item We introduce a simple yet effective method inspired by mixup, which encodes 3D patches of a target image by mixing them to reference patches with known ground-truth. Unlike FL approaches, which require a bulky training setup, or homomorphic networks which are computationally prohibitive, our method works in a normal training setup and has a low computational overhead.

\item We also propose a learning approach for recovering the target segmentation maps from mixed ones, which improves the noisy results of directly reversing the mixing function. 

\item Results are further improved with a test-time augmentation strategy that mixes a target image with different references and then ensembles the segmentation predictions to achieve a higher accuracy.  

\item We conduct extensive experiments on two challenging 3D brain MRI benchmarks, and show our method to largely outperform state-of-art approaches for privacy-preserving segmentation, while being simpler and faster than these approaches and yet offering a similar level of privacy.
\end{enumerate}



\vspace{-0.25cm}
\section{Related works}
\label{sec:rw}
\vspace{-0.25cm}



Most privacy-preserving approaches for image analysis fall in two categories: those based on homomorphic encryption and the ones using adversarial learning.

\mypar{Homomorphic encryption (HE)~\cite{Dowlin16,Hesamifard17,nandakumar2019towards}} This type of encryption enables to compute a given function on encrypted data without having to decrypt it first or having access to the private key. Although HE offers strong guarantees on the security of the encrypted data, this approach suffers from two important limitations: 1) it has a prohibitive computational/communication overhead~\cite{rouhani2018}; 2) it is limited to multiplications and additions, and non-linear activation functions have to be approximated by polynomial functions. 
As a result, homomorphic networks have been relatively simplistic~\cite{Hardy2017}, and even computing the output of a simple CNN is prohibitively slow (e.g., 30 minutes for a single image~\cite{nandakumar2019towards}).


\mypar{Adversarial learning (AL)} This type of approach uses a neural net to \change{encode images} so that private information is discarded, yet the encoded image still holds the necessary information to perform a given image analysis task~\cite{xu2019ganobfuscator,raval2017protecting}. The encoder is trained jointly with two downstream networks taking the encoded image as input, the first one seeking to perform the target task and the other one (the discriminator) trying to recover the private information. The parameters of the encoder are \change{updated to minimize} the task-specific utility loss while maximizing the loss of the discriminator. In medical imaging tasks, where patient identity should be protected, the discriminator cannot be modeled as a standard classifier since the number of classes (e.g., patient IDs) is not fixed. To alleviate this problem, the method in~\cite{kim2020privacynet} uses a Siamese discriminator which receives two encoded images as input and predicts if the images are from the same patient or not. While input images are encoded, the method produces non-encoded segmentation maps which  may still be used to identify the patient. The authors of \cite{bach2021nonlinear} overcome this limitation by transforming input images with a reversible non-linear warping which depends on a private key. When receiving a deformed segmentation map from the server, the client can recover the true segmentation by reversing the transformation. However, as the method in \cite{kim2020privacynet}, this approach requires multiple scans of the same patient to train the Siamese discriminator, which may not be available in practice. Furthermore, the learned encoder is highly sensitive to the distribution of input images and fails to obfuscate identity when this distribution shifts. In contrast, our method does not require multiple scans per patient. It is also simpler to train and, because it relies on the general principle of BSS, is less sensitive to the input image distribution.

\vspace{-0.25cm}
\section{Methodology}
\vspace{-0.25cm}

We first \change{introduce} the principles of blind source separation and mixup on which our work is based, and then present the details of our Mixup-Privacy method.   

\vspace{-0.25cm}
\subsection{Blind source separation} 
\label{section:BSS}
\vspace{-0.25cm}

Blind source separation (BSS) is a well-known problem of signal processing which seeks to recover a set of unknown source signals from a set of mixed ones, without information about the  mixing process. Formally, let $x(t) = [x_1(t), \ldots, x_n(t)]^T$ be a set of $n$ source signals which are mixed into a set of $m$ signals, $y(t) = [y_1(t), \ldots, y_m(t)]^T$, using matrix $A \in \mathbb{R}^{m \times n}$ as follows: $y(t) = A\!\cdot\!x(t)$. BSS can be defined as recovering $x(t)$ when given only $y(t)$. While efficient methods exist for cases where $m=n$, the problem is much harder to solve when $m<n$ as the system of equations then becomes under-determined~\cite{article_BSS}. For the extreme case of single channel separation ($n\!=\!1$), \cite{DAVIES20071819} showed that traditional approaches such as Independent Component Analysis (ICA) fail when the sources have substantially overlapping spectra. Recently, the authors of \cite{DBLP:journals/corr/abs-2002-07942} proposed a deep learning method for single channel separation, using the noise-annealed Langevin dynamics to sample from the posterior distribution of sources given a mixture. Although it achieves impressive results for the separation of RGB natural images, as we show in our experiments, this method does not work on low-contrast intensity images such as brain MRI. Leveraging the ill-posed nature of single source separation, we encode 3D patches of images to segment by mixing them with those of reference images.   

\vspace{-0.2cm}
\subsection{Mixup training} \label{section:mix_up}
\vspace{-0.1cm}

Mixup is a data augmentation technique that generates new samples via linear interpolation between random pairs of images as well as their associated one-hot encoded labels~\cite{DBLP:journals/corr/abs-1710-09412}. Let $(x_i, y_i)$ and $(x_j, y_j)$ be two examples drawn at random from the training data, and $\alpha \sim \mathrm{Beta}(b,b)$ be a mixing coefficient sampled from the Beta distribution with hyperparameter $b$. Mixup generates virtual training examples $(\tilde{x},\tilde{y})$ as follows: 
\begin{equation}
\tilde{x} \, = \, \alpha\sskip x_i \, + \, (1\!-\!\alpha)\sskip x_j; \quad \tilde{y} \,=\, \alpha\sskip y_i \, + \, (1\!-\!\alpha)\sskip y_j.
\end{equation}
While Mixup training has been shown to bring  performance gains in various problems, including image classification \cite{guo2019mixup} and semantic segmentation \cite{zhou2022generalizable}, it has not been explored as a way to preserve privacy in medical image segmentation.

\begin{figure}[ht!]
    \centering
    \includegraphics[width=1\textwidth]{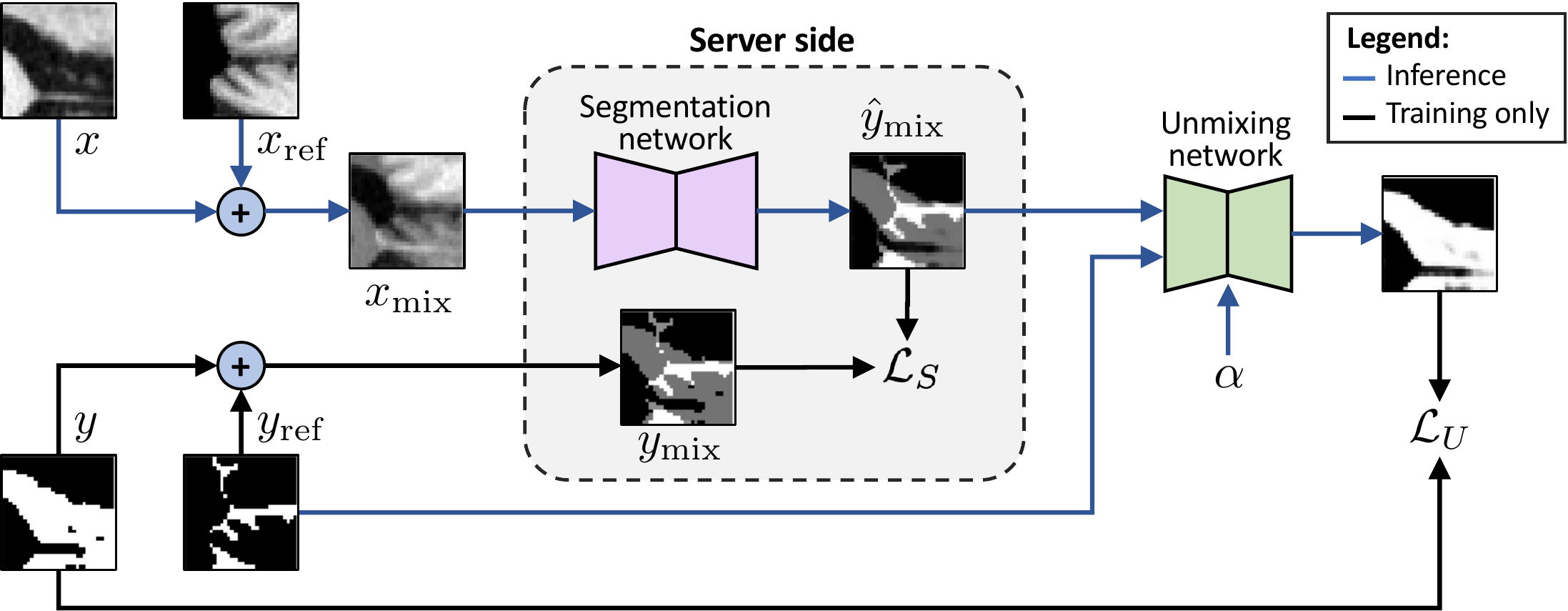}
    \caption{Training diagram of the proposed system with the client (left and right) and the server (middle).  The client mixes the input image $x$ and segmentation map $y$ with a reference pair $(x_{\mr{ref}},y_{\mr{ref}})$. The mixed data is then fed to a segmentation network located on a server and whose output is a mixed segmentation map. The resulting segmentation map is sent back to the client, which decodes it with a unmixing network and the reference map $y_{\mr{ref}}$.\vspace{-0.4cm} }
    \label{fig:system_diagram}
\end{figure}

\vspace{-.1cm}
\subsection{Proposed system}
As shown in Fig~\ref{fig:system_diagram}, our method involves a client which has an image $x$ to segment and a server which has to perform segmentation without being able to recover private information from $x$. During training, the client mixes an image $x$ and its associated segmentation map $y$ with a reference data pair $x_{\mr{ref}}$ and $y_{\mr{ref}}$. The mixed data $(x_{\mr{mix}},y_{\mr{mix}})$ is then sent to the server. Since unmixing images requires to solve an under-determined BSS problem, $x$ cannot be recovered from $x_{\mr{mix}}$ without $x_{\mr{ref}}$. This renders $x_{\mr{mix}}$ unusable if intercepted by an unauthorized user. During inference, the server network returns the mixed segmentation maps $\hat y_{\mr{mix}}$ to the client, which then recovers the true segmentation maps $y$ by reversing the mixing process. The individual steps of our method, which is trained end-to-end, are detailed below.

\mypar{Data mixing.} Since 3D MR images are memory heavy, our segmentation method processes images in a patch-wise manner. Each patch $x \in \mathbb{R}^{H\times W\times D}$ is mixed with a reference patch of the same size: 
\begin{align}
    x_{\mr{mix}} \, = \, \alpha\sskip x_{\mr{target}} \, + \, (1\!-\!\alpha) \sskip x_{\mr{ref}},
    \label{eq:mix_image}
\end{align}
where $\alpha \in [0,1]$ is a mixing weight drawn randomly from the uniform distribution\footnote{Unlike Mixup which uses the Beta distribution to have a mixing weight close to 0 or 1, we use the uniform distribution to have a broader range of values.}. During training, the one-hot encoded segmentation ground-truths $y\in[0,1]^{C\times H\times W\times H}$ are also mixed using the same process:
\begin{align}
    y_{\mr{mix}} \, = \, \alpha \sskip y_{\mr{target}} \, + \, (1\!-\!\alpha) \sskip y_{\mr{ref}}
    \label{eq:mix_segmentation_gt},
\end{align}
and are sent to the server with the corresponding mixed image patches $x_{\mr{mix}}$.
 
\mypar{Segmentation and unmixing process.} The server-side segmentation network $S(\cdot)$ receives a mixed image patch $x_{\mr{mix}}$, predicts the mixed segmentation maps $\hat y_{\mr{mix}} = S(x_{\mr{mix}})$ as in standard Mixup training, and then sends $\hat y_{\mr{mix}}$ back to the client. Since the client knows the ground-truth segmentation of the reference patch, $y_{\mr{ref}}$, it can easily recover the target segmentation map by reversing the mixing process as follows: 
\begin{equation}
\label{eq:naive_rec}
\hat{y}_{\mr{target}} \, = \, \frac{1}{\alpha}\big(\hat{y}_{\mr{mix}} \, -  \,(1\!-\!\alpha)y_{\mr{ref}}\big).
\end{equation}
However, since segmenting a mixed image is more challenging than segmenting the ones used for mixing, the naive unmixing approach of Eq. (\ref{eq:naive_rec}) is often noisy. 
\change{To address this problem}, we use a shallow network $D(\cdot)$ on the client side to perform this operation. Specifically, this unmixing network receives as input the mixed segmentation $\hat{y}_{\mr{mix}}$, the reference segmentation $y_{\mr{ref}}$, and the mixing coefficient $\alpha$, and predicts the target segmentation as $\hat y_{\mr{target}}= D(\hat{y}_{\mr{mix}},y_{\mr{ref}},\alpha)$.

\vspace{-.1cm}
\subsection{Test-time augmentation}\label{sec:TTA}

Test-time augmentation (TTA) is a simple but powerful technique to improve performance during inference~\cite{wang2019aleatoric}. Typical TTA approaches generate multiple augmented versions of an example $x$ using a given set of transformations, and then \change{combine the} predictions for these augmented examples based on an ensembling strategy. In this work, we propose a novel TTA approach which augments a target patch $x_{\mr{target}}$ by mixing it with different reference patches $\{x_{\mr{ref}}^k\}_{k=1}^K$:
\begin{equation}
x_{\mr{mix}}^k \, = \, \alpha \sskip x_{\mr{target}} \, + \, (1\!-\!\alpha) \sskip x_{\mr{ref}}^k.
\end{equation}
The final prediction for the target segmentation is then obtained by averaging the \change{predictions} of individual mixed patches:
\begin{equation}
\label{eq:TAA}
\hat{y}_{\mr{target}} \, = \, \frac{1}{K} \sum_{k=1}^K D\big(\hat y^k_{\mr{mix}},\, y_{\mr{ref}}^k, \,\alpha\big).
\end{equation}
As we will show in experiments, segmentation accuracy can be significantly boosted using only a few augmentations.

\section{Experimental setup}
\label{sec:exp}
\vspace{-.5em}


\mypar{Datasets.} We evaluate our method on the privacy-preserving segmentation of brain MRI from two public benchmarks, the Parkinson’s Progression Marker Initiative (PPMI) dataset~\cite{marek2011ppmi} and the Brain Tumor Segmentation (BraTS) 2021 Challenge dataset. For the PPMI dataset, we used T1 images from 350 subjects for segmenting brain images into three tissue classes: white matter (WM), gray matter (GM) and cerebrospinal fluid (CSF). 
Each subject underwent one or two baseline acquisitions and one or two acquisitions 12 months later for a total of 773 images. The images were registered onto a common MNI space and resized to $144\times192\times160$ with a 1mm$^3$ isotropic resolution. We divided the dataset into training and testing sets containing 592 and 181 images, respectively, so that images from the same subject are not included in both the training and testing sets. Since PPMI has no ground-truth annotations, as in \cite{kim2020privacynet,bach2021nonlinear}, we employed Freesurfer to obtain a pseudo ground-truth for training. We included the PPMI dataset in our experiments because it has multiple scans per patient, which is required for some of the compared baselines \cite{kim2020privacynet,bach2021nonlinear}.

BraTS 2021 is the largest publicly-available and fully-annotated dataset for brain tumor segmentation. It contains 1,251 multi-modal MRIs of size $240\!\times\!240\!\times\!155$. 
Each image was manually annotated with four labels: necrose (NCR), edema (ED), enhance tumor (ET), and background.  We excluded the T1, T2 and FLAIR modalities and only use T1\textsc{ce}. From the 1,251 scans, 251 scans were used for testing, while the remaining constituted the training set. 

\mypar{Evaluation metrics.} Our study uses the 3D Dice similarity coefficient (DSC) to evaluate the segmentation performance of tested methods. For measuring the ability to recover source images, we measure the Multi-scale Structural Similarity (MS-SSIM)~\cite{wang2003multiscale} between the original source image and the one recovered from a BSS algorithm~\cite{DBLP:journals/corr/abs-2002-07942}. Last, to evaluate the privacy-preserving ability of our system, we model the task of recovering a patient's identity as a retrieval problem and measure performance using the standard F1-score and mean average precision (mAP) metrics. 

\mypar{Implementation details.} We used patches of size $32\!\times\!32\!\times\!32$ for PPMI and $64\!\times\!64\!\times\!64$ for BraTS. Larger patches were considered for BraTS to capture the whole tumor. We adopted architectures based on U-Net~\cite{DBLP:journals/corr/RonnebergerFB15} for both the segmentation and unmixing networks. For the more complex segmentation task, we used the U-Net++ architecture described in \cite{10.1007/978-3-030-00889-5_1}, whereas a small U-Net with four \change{convolutional} blocks was employed for the unmixing network. For the latter, batch normalization layers were replaced by adaptive instance normalization layers~\cite{huang2017arbitrary} which are conditioned on the mixing coefficient $\alpha$. Both the segmentation and unmixing networks are trained using combination of multi-class cross entropy loss and 3D Dice loss~\cite{milletari2016v}. End-to-end training was performed for 200,000 iterations on a NVIDIA A6000 GPU, using the Adam optimizer with a learning rate of $1\times10^{\!-\!4}$ and a batch size of 4.

\mypar{Compared methods.} We evaluate different variants of our Mixup-Privacy method for privacy-preserving segmentation. For the segmentation unmixing process, two approaches were considered: a \emph{Naive} approach which reverses the mixing process using Eq. (\ref{eq:naive_rec}), and a \emph{Learned} one using the unmixing network $D(\cdot)$. Both approaches were tested with and without the TTA strategy described in Section \ref{sec:TTA}, giving rise to four different variants. We compared these variants against a segmentation \emph{Baseline} using non-encoded images and two recent approaches for cloud-based privacy-preserving segmentation: \emph{Privacy-Net}~\cite{kim2020privacynet} and \emph{Deformation-Proxy}~\cite{bach2021nonlinear}. The hyperparameters of all compared methods were selected using 3-fold cross-validation on the training set.


  \begin{table}[t!]
    \centering
    \caption{Main results of the proposed approach across different tasks \change{- including segmentation, blind source separation and test-retest reliability -} and two datasets (PPMI and BraTS2021).}
    \setlength{\tabcolsep}{2pt}
    \label{table:experiment_result}
    \begin{footnotesize}
        \begin{tabular}{lccccccccc}
        \toprule
        & \multicolumn{4}{c}{PPMI} & &\multicolumn{4}{c}{BraTS2021} \\
        \cmidrule(l{5pt}r{5pt}){2-5} 
        \cmidrule(l{5pt}r{5pt}){7-10} 
        & GM & WM & CSF & Avg & \phantom{M}
        & NCR & ED & ET & Avg\\
        \midrule
        \rowcolor{Gray}
        \multicolumn{10}{c}{\textsc{Segmentation \change{(Dice Score)}}}\\
        \midrule    
        Baseline & 0.930 & 0.881 & 0.876 & 0.896 & & 0.846 & 0.802 & 0.894 & 0.847  \\
        \midrule
        Privacy-Net~\cite{kim2020privacynet} & 0.905 & 0.804 & 0.732 & 0.813 & & --- & --- & --- & --- \\
        Deformation-Proxy~\cite{bach2021nonlinear}~~ & 0.889 & 0.825 & 0.757 & 0.823 &  & --- & --- & --- & --- \\
        Ours (\emph{Naive}) & 0.758 & 0.687 & 0.634 & 0.693 & & 0.656 & 0.635 & 0.692 & 0.661 \\
        Ours (\emph{Naive\,+\,TTA}) & 0.852 & 0.829 & 0.793 & 0.825 & & 0.775 & 0.737 & 0.804 & 0.772 \\
        Ours (\emph{Learned}) & 0.893 & 0.833 & 0.795 & 0.840 & & 0.805 & 0.763 & 0.842 & 0.803 \\
        Ours (\emph{Learned\,+\,TTA}) & \bf 0.925 & \bf 0.879 & \bf 0.863 & \bf 0.889 & & \bf 0.841 & \bf 0.808 &\bf  0.872 & \bf 0.840 \\
        \midrule
        \rowcolor{Gray}
        \multicolumn{10}{c}{\textsc{Blind Source Separation \change{(MS-SSIM)}}}\\
        \midrule
        Separation Accuracy & \multicolumn{4}{c}{$0.602\pm0.104$} & &\multicolumn{4}{c}{$0.588\pm0.127$} \\
        \midrule        
        \rowcolor{Gray}
        \multicolumn{10}{c}{\textsc{Test-Retest Reliability \change{(ICC value)}}}\\
        \midrule
        ICC & 0.845 & 0.812 & 0.803 & --- & & 0.842 & 0.812  & 0.803 & --- \\
        Upper bound & 0.881 & 0.856 & 0.844 & --- & & 0.878 & 0.855 & 0.839 & --- \\
        Lower bound & 0.798 & 0.783 & 0.771 & --- & & 0.805 & 0.777 & 0.768 & --- \\        
        \bottomrule
        \end{tabular}
        \end{footnotesize}
    \end{table}

\vspace{-0.3cm}
\subsection{Results}    
\vspace{-0.25cm}
    
\begin{figure}[!ht]
        \centering        
        \begin{footnotesize}
        
        \setlength{\tabcolsep}{6pt}
        \begin{tabular}{c|ccccc}            
            \includegraphics[width=.12\textwidth]{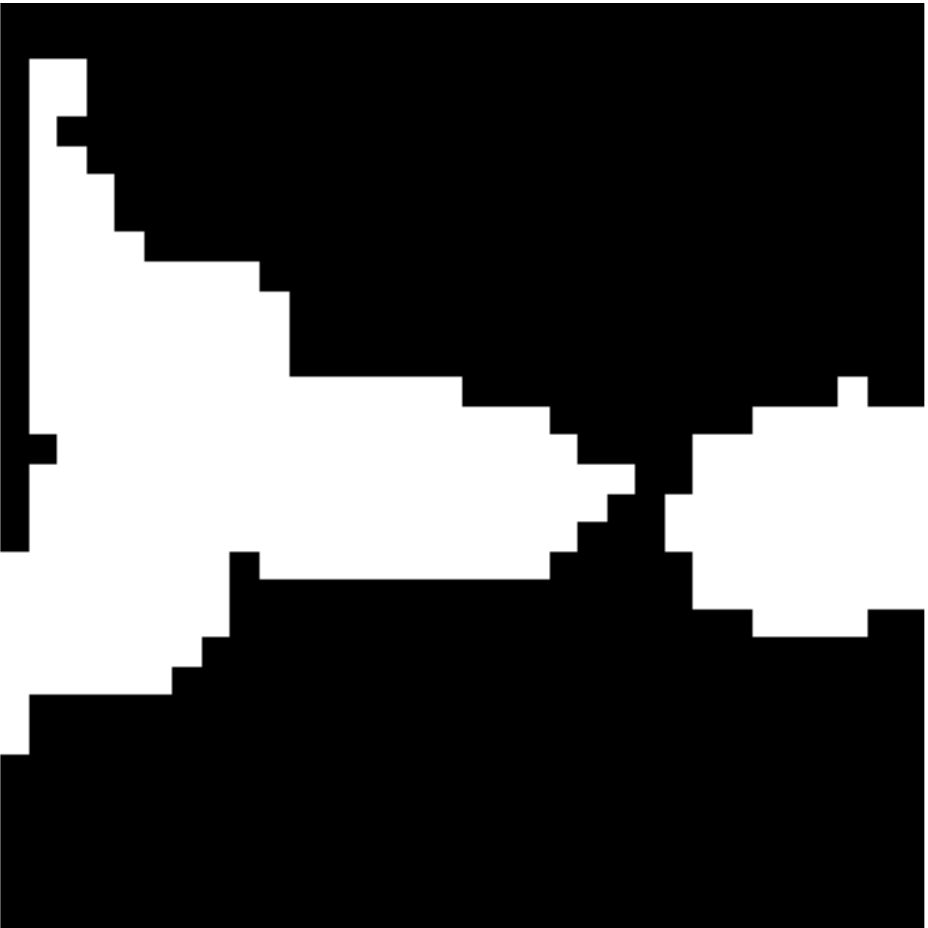}
            &\includegraphics[width=.12\textwidth]{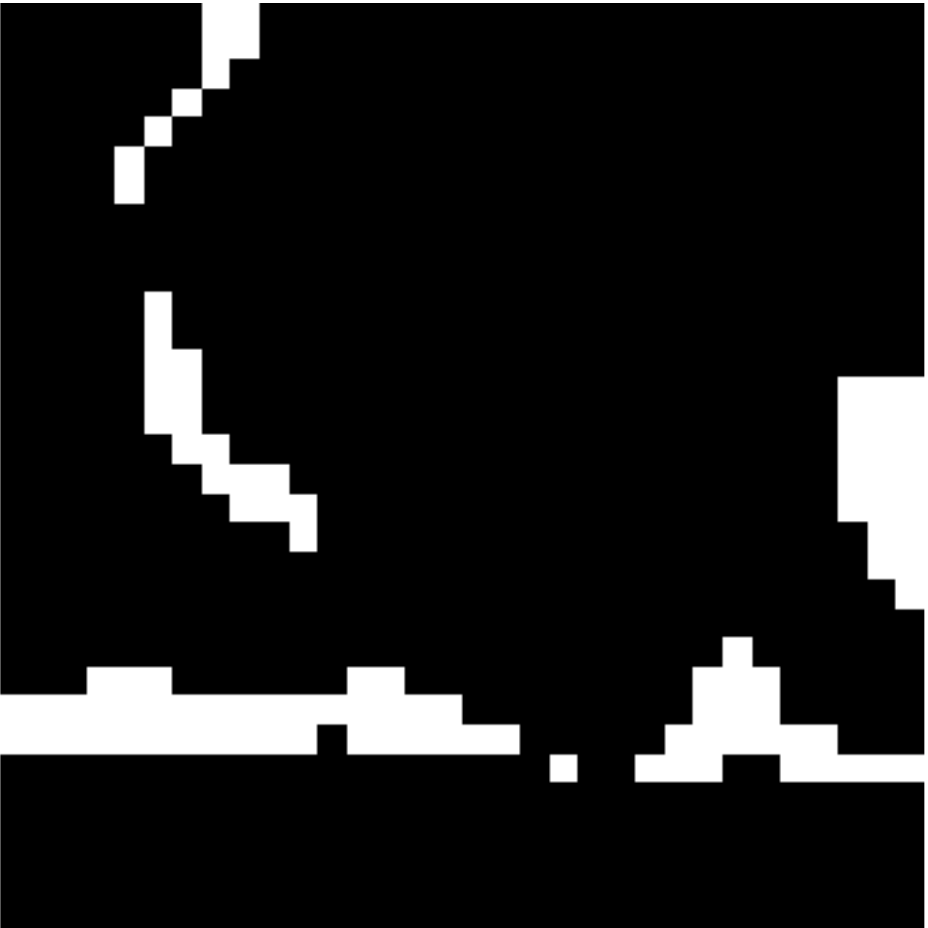}
            &\includegraphics[width=.12\textwidth]{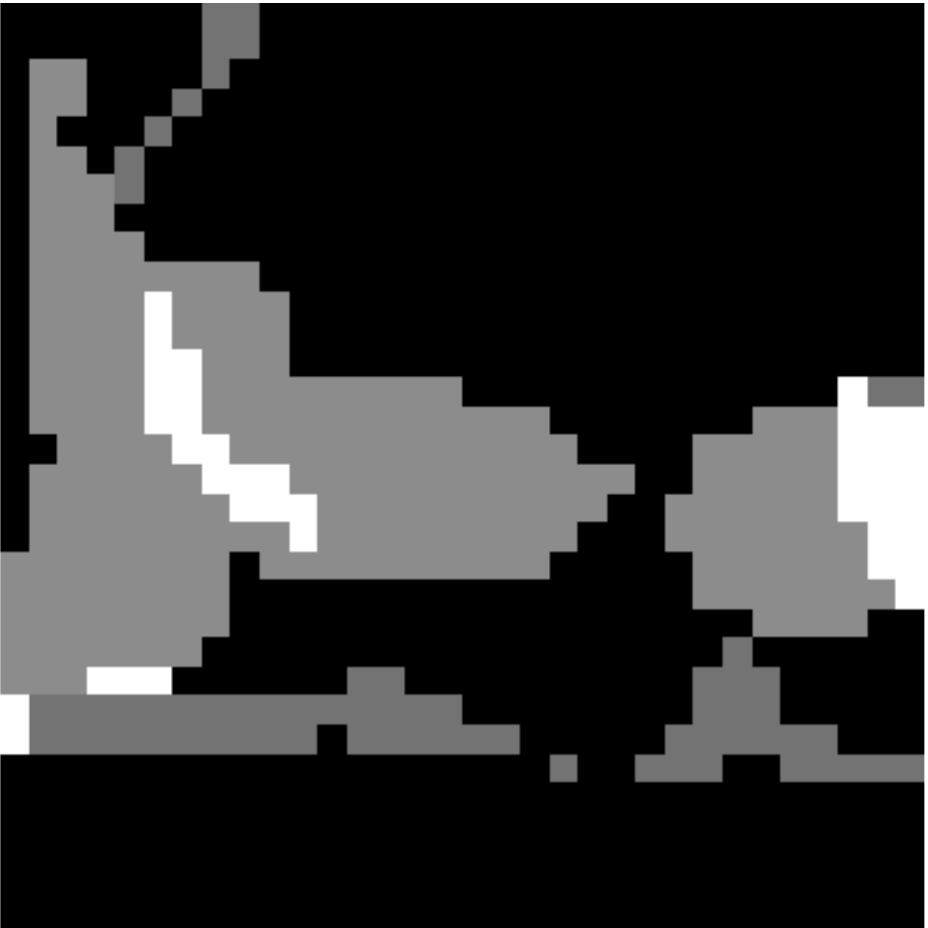}
            &\includegraphics[width=.12\textwidth]{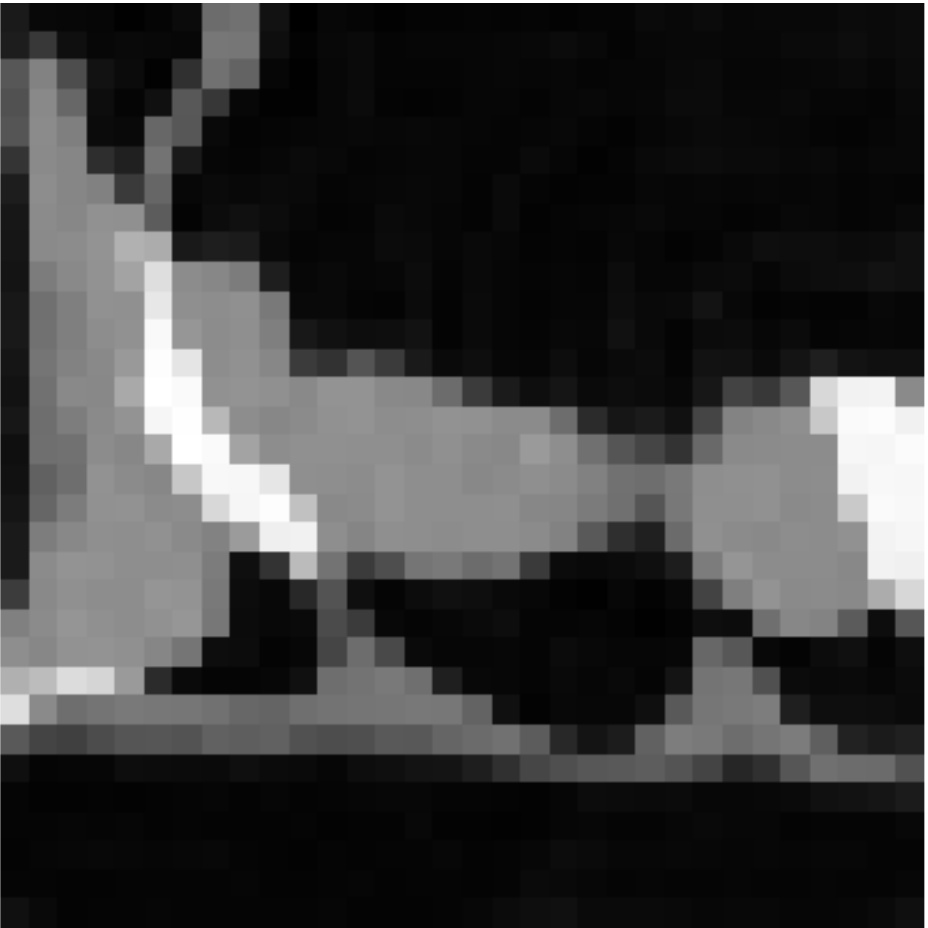} 
            &\includegraphics[width=.12\textwidth]{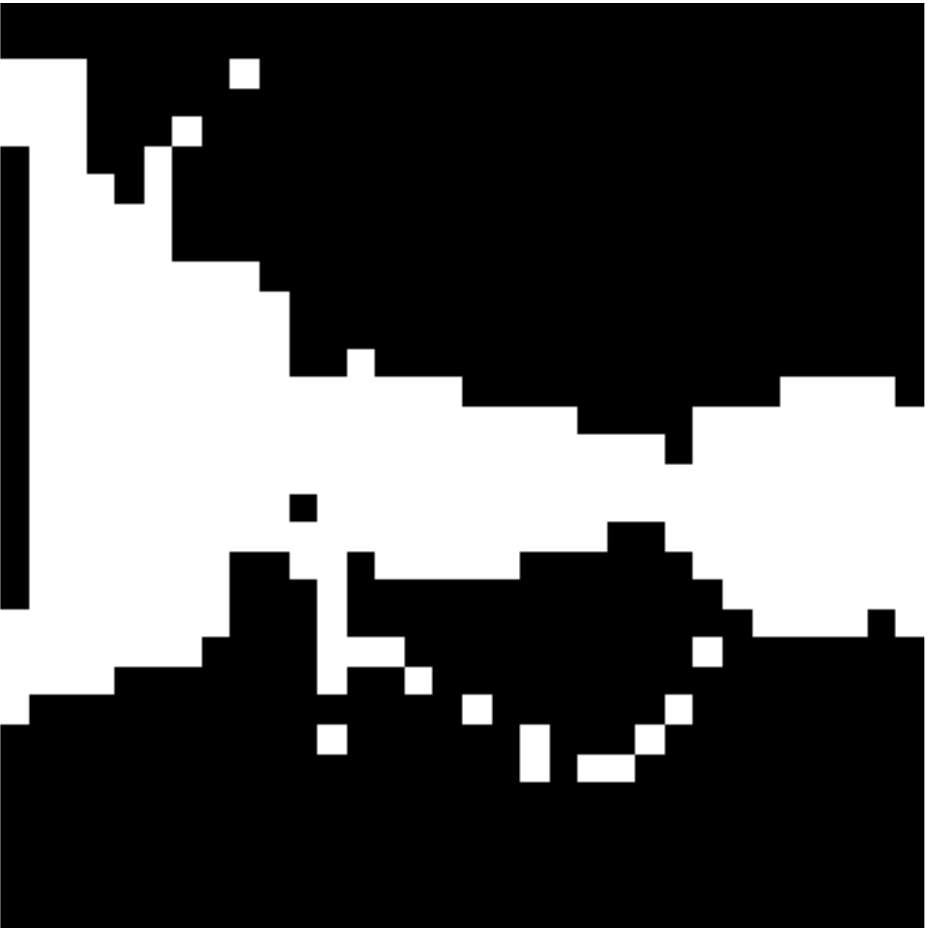}
            &\includegraphics[width=.12\textwidth]{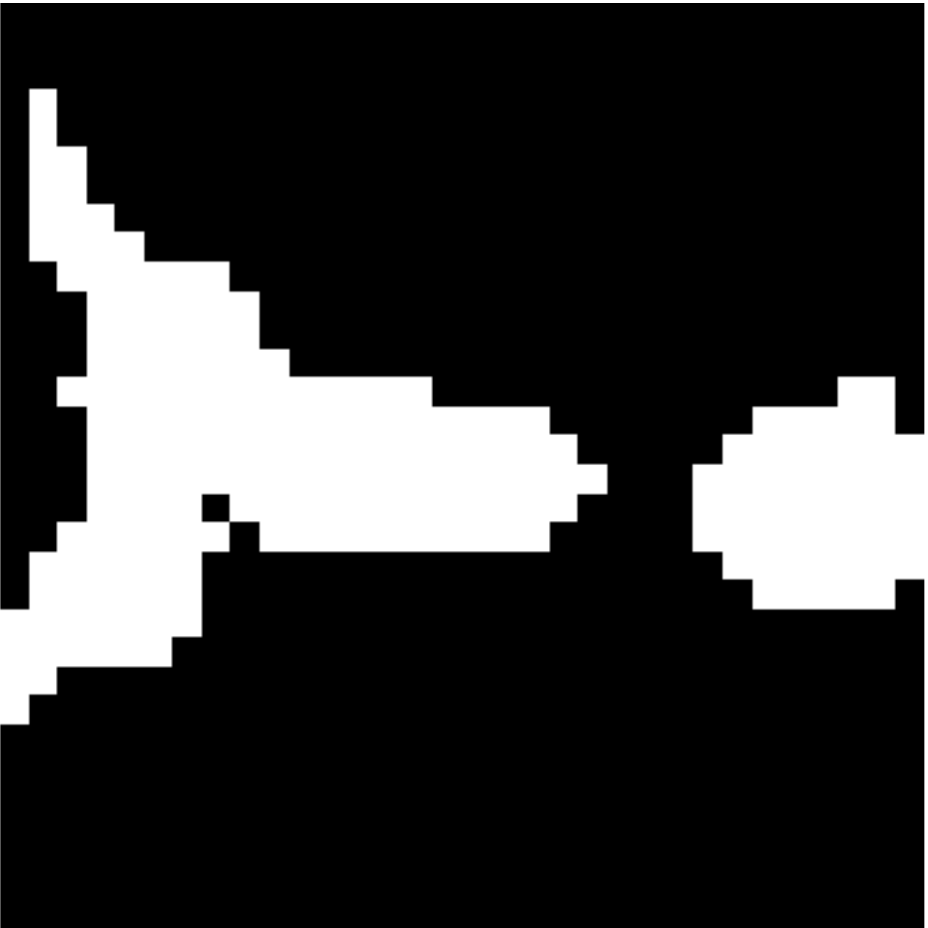}\\
    \bf Target & Reference 1 & Mixed GT  & Prediction & Naive & Learned\\[.4em]
            \includegraphics[width=.12\textwidth]{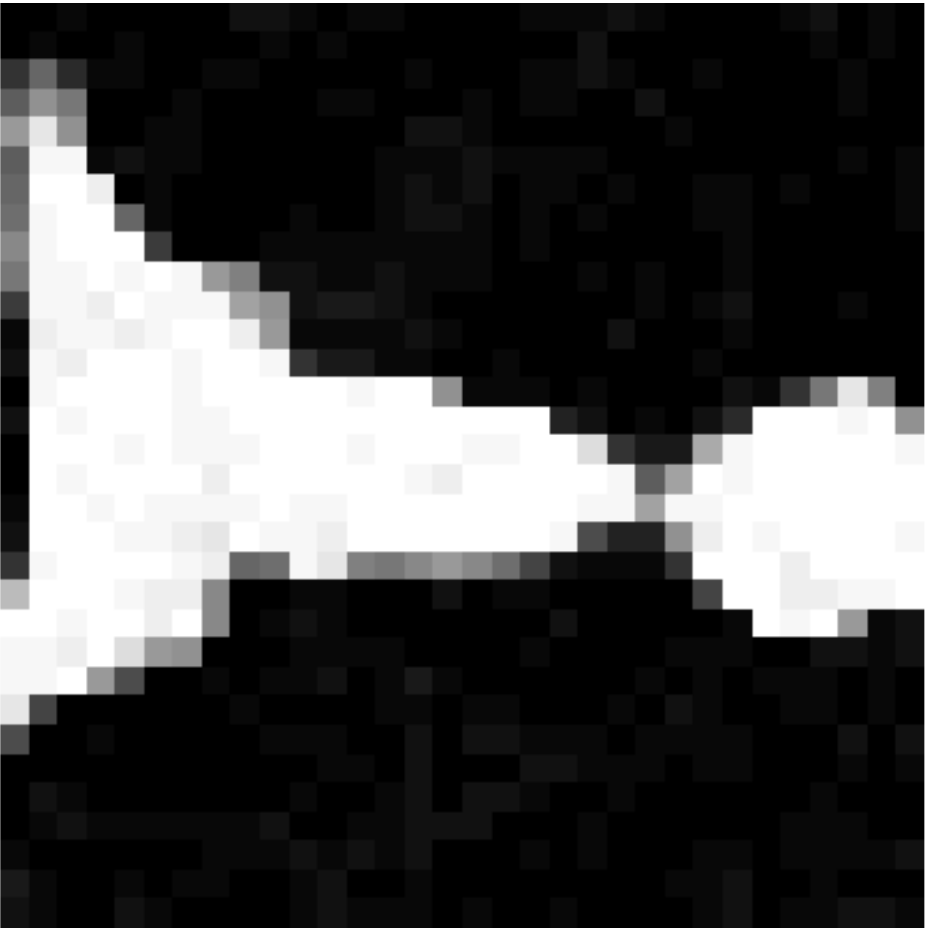}
                    
            &\includegraphics[width=.12\textwidth]{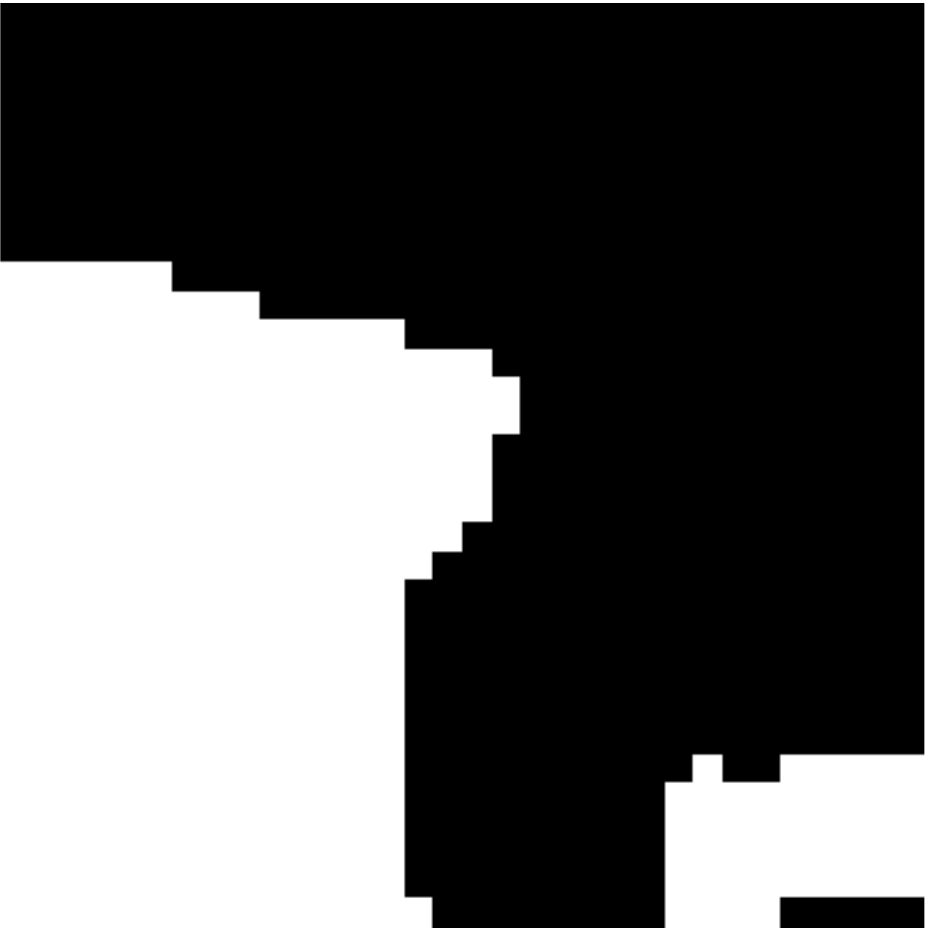}
            &\includegraphics[width=.12\textwidth]{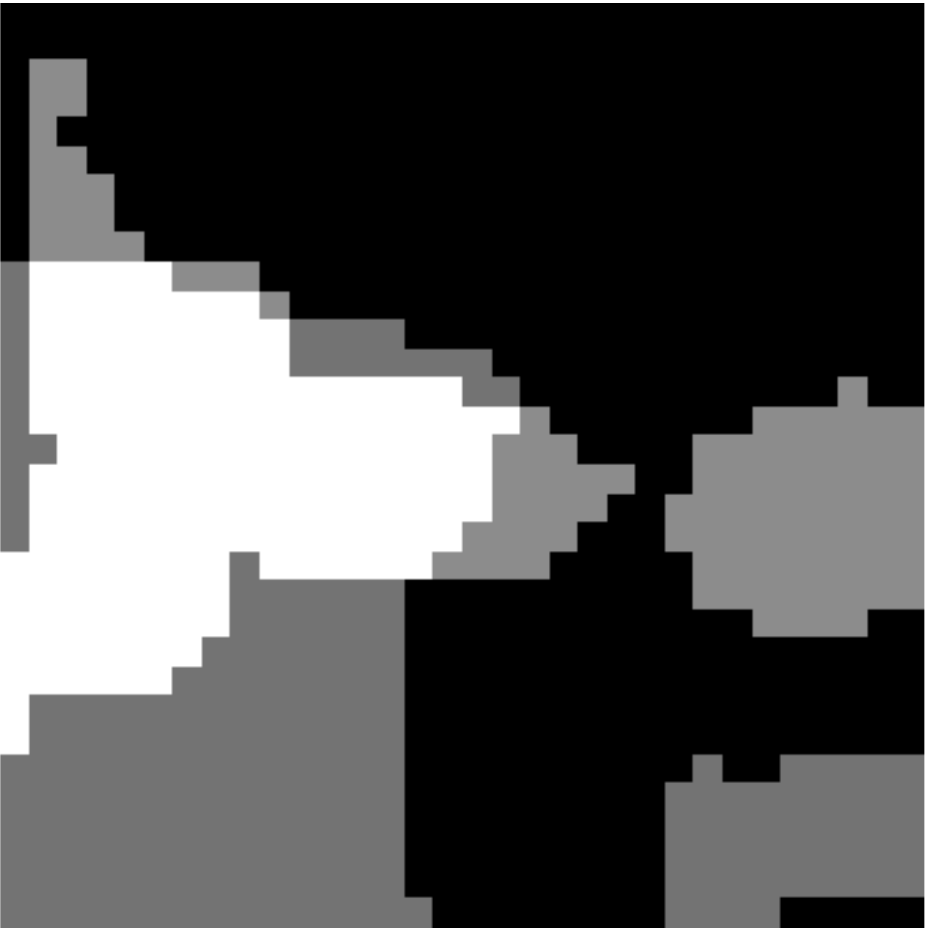}
            &\includegraphics[width=.12\textwidth]{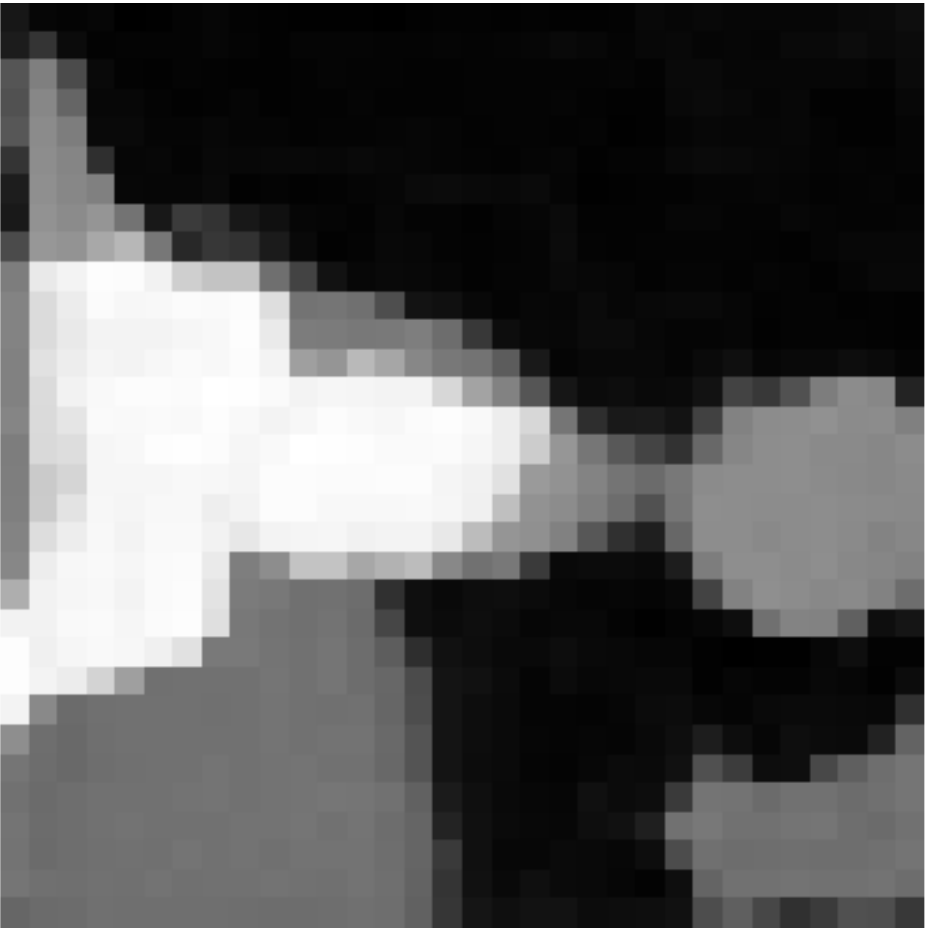} 
            &\includegraphics[width=.12\textwidth]{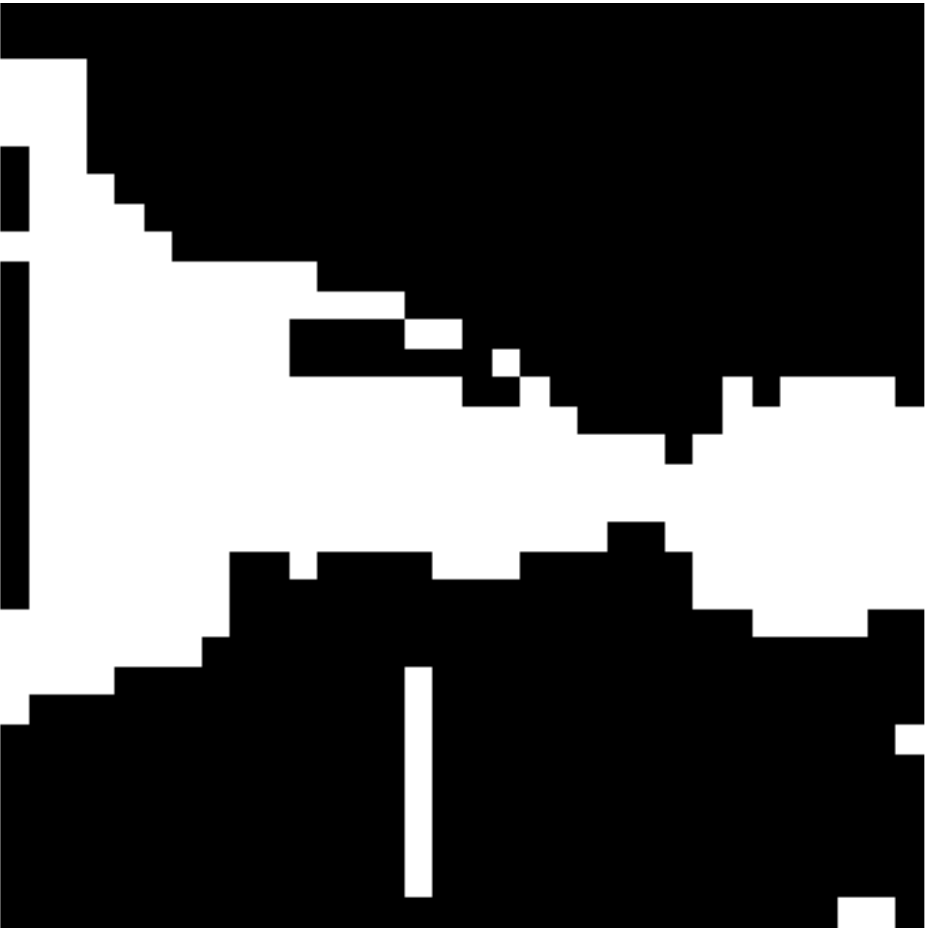}
            &\includegraphics[width=.12\textwidth]{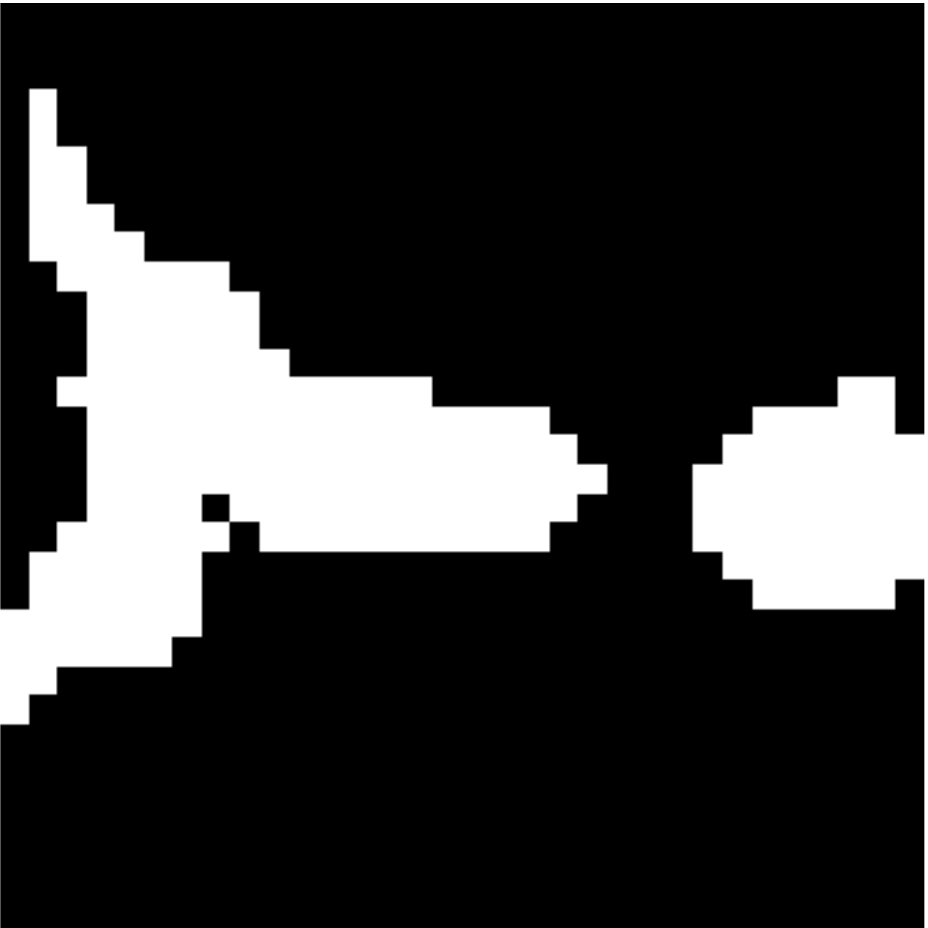}\\
            Naive\,+\,TTA & Reference 2 & Mixed GT  & Prediction & Naive & Learned\\[.4em]
            \includegraphics[width=.12\textwidth]{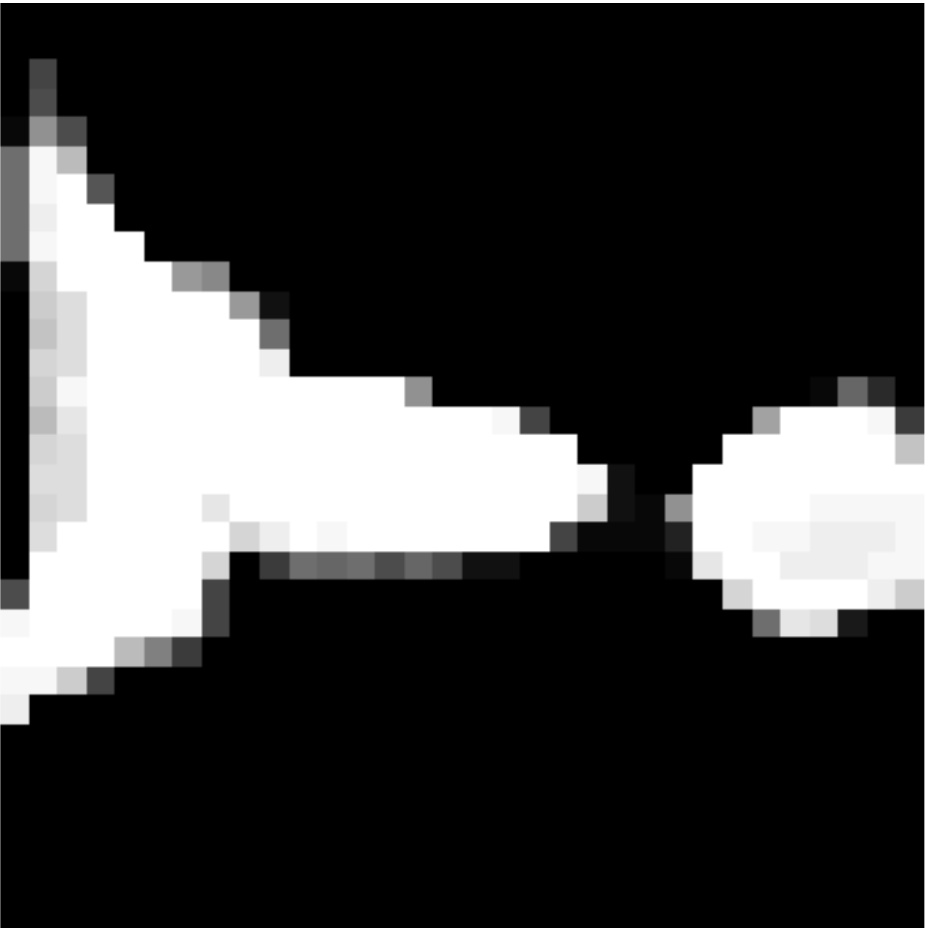}
            &\includegraphics[width=.12\textwidth]{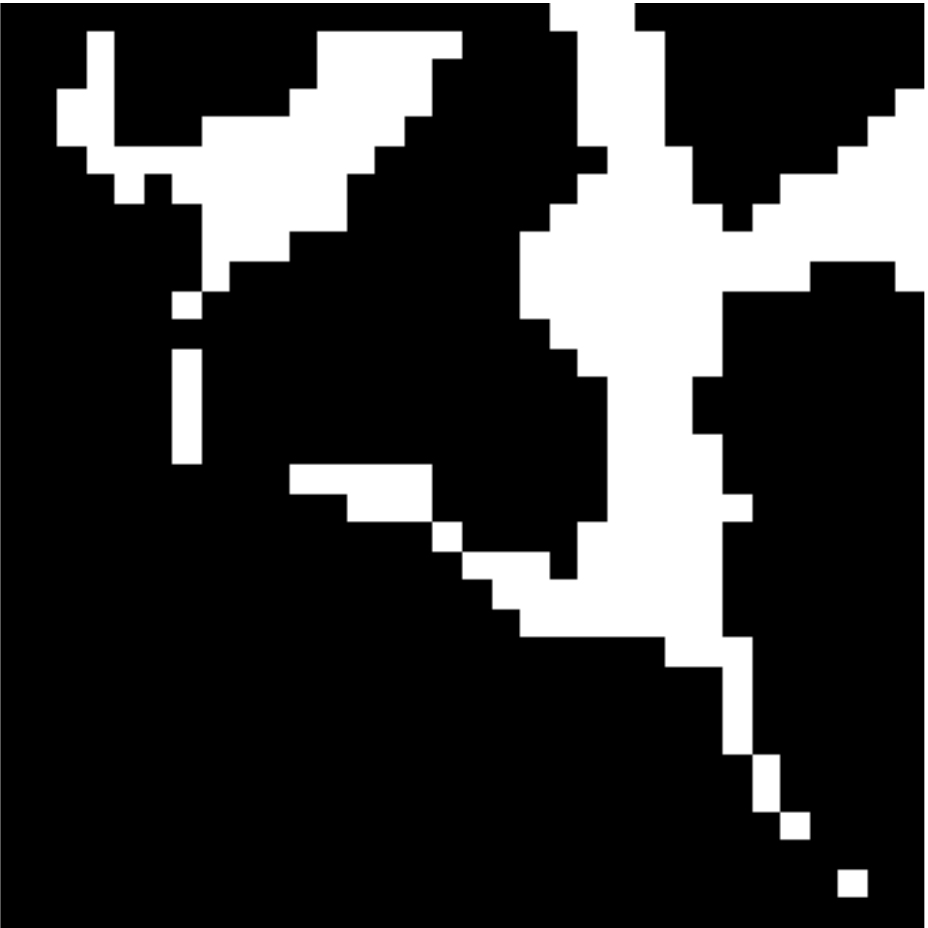}
            &\includegraphics[width=.12\textwidth]{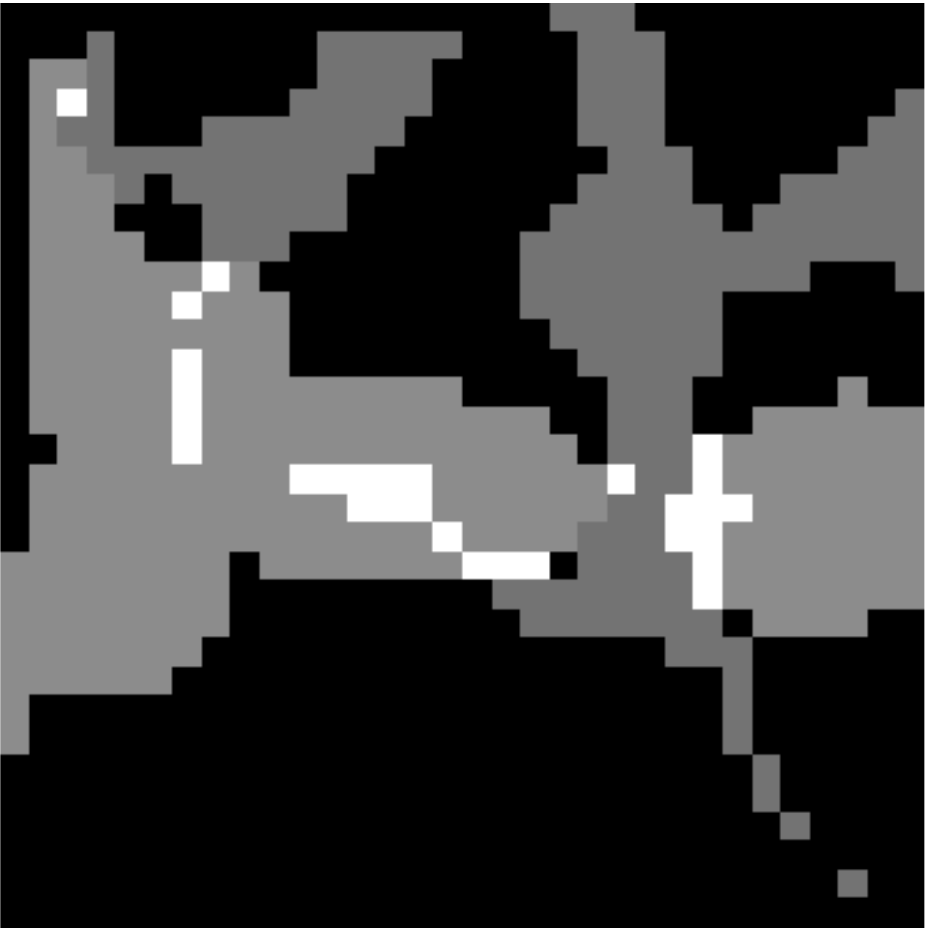}
            &\includegraphics[width=.12\textwidth]{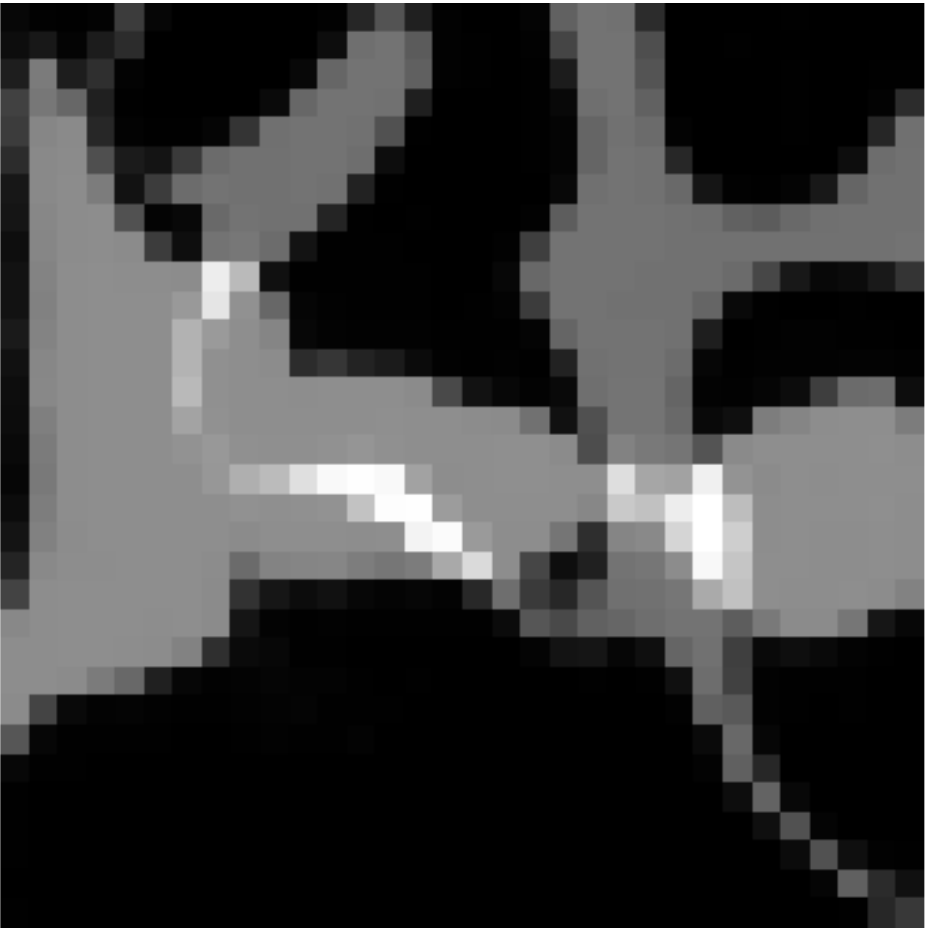} 
            &\includegraphics[width=.12\textwidth]{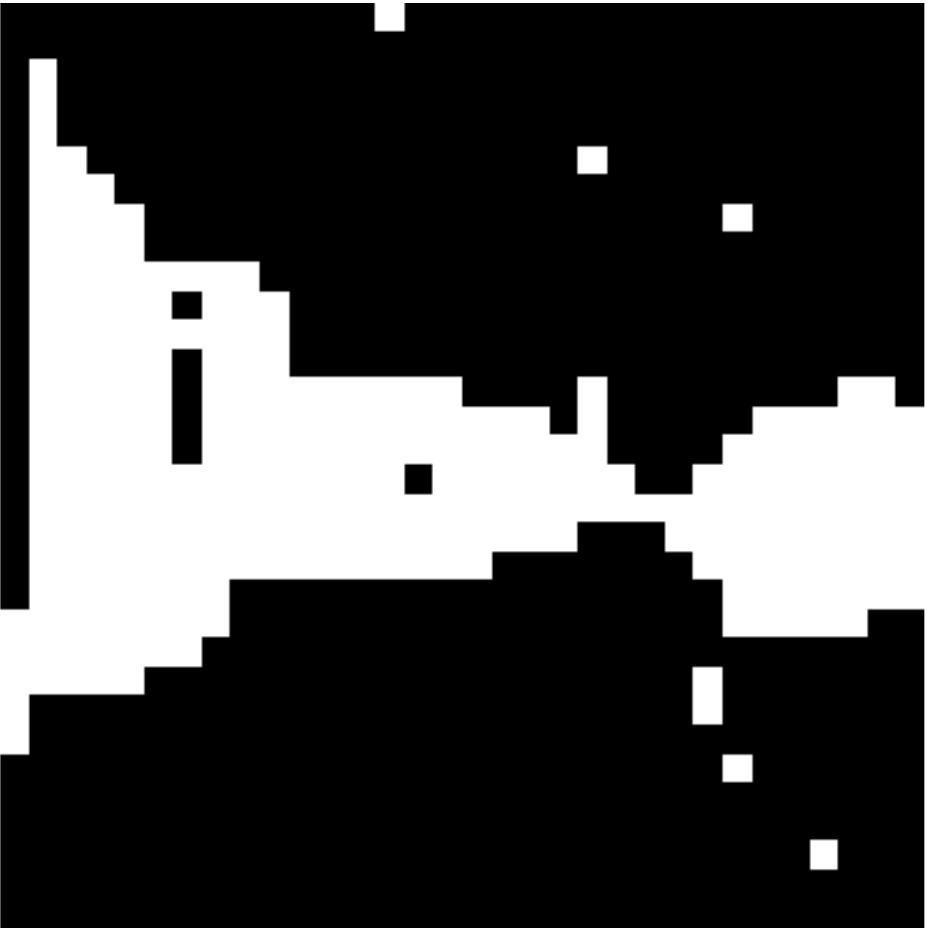}
            &\includegraphics[width=.12\textwidth]{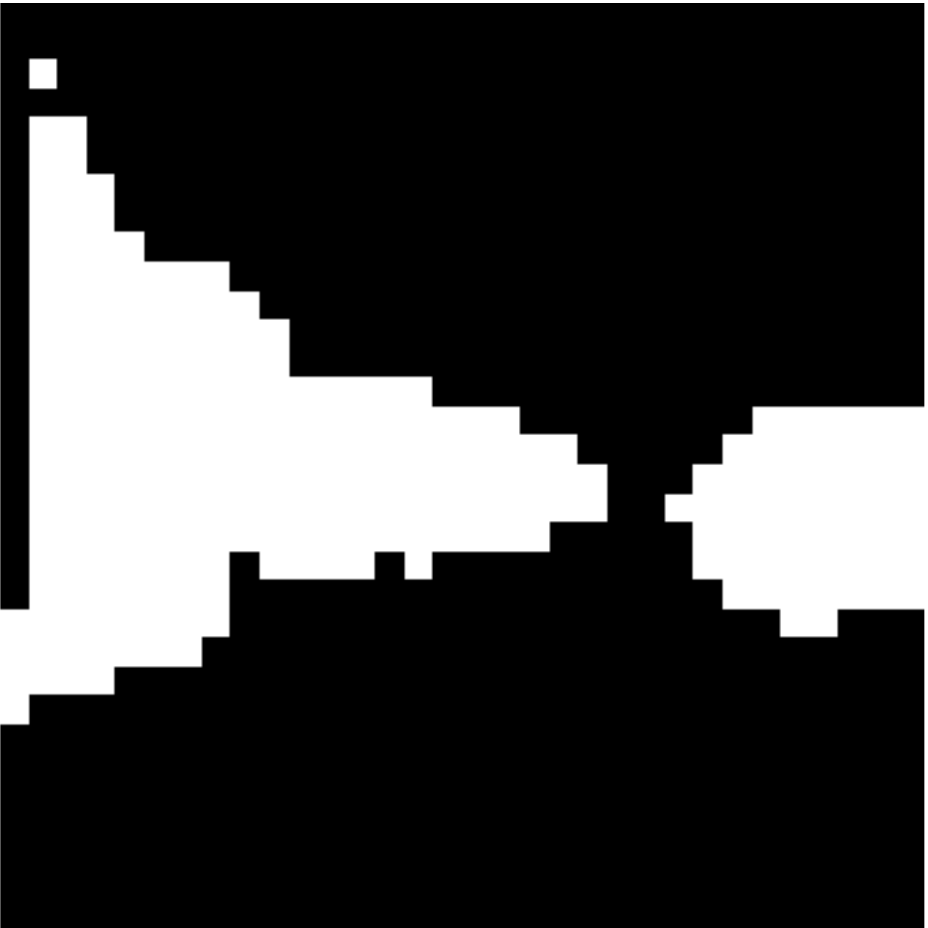}\\
            Learned\,+\,TTA & Reference 3 & Mixed GT  & Prediction & Naive & Learned\\[-.5em]
        \end{tabular} 
        \end{footnotesize}
        \caption{Examples of segmented patches obtained by the naive and learned unmixing approaches from the same target and three different references. Naive\,+\,TTA and Learned\,+\,TTA show the mean prediction of these approaches for 30 augmentations (each one using a different reference).}
        \vspace{-0.4cm}
        \label{fig:segmenation_patch}
        \end{figure}

\mypar{Segmentation performance.} The top section of Table \ref{table:experiment_result} reports the segmentation performance of the compared models. Since Privacy-Net and Deformation-Proxy require longitudinal data to train the Siamese discriminator, we only report their results for PPMI, which has such data. Comparing the naive and learned approaches for segmentation unmixing, we see that using an unmixing network brings a large boost in accuracy. Without TTA, the learned unmixing yields an overall Dice improvement of 14.7\% for PPMI and of 14.2\% for BraTS2021. As shown in Fig.~\ref{fig:segmenation_patch}, the naive approach directly reversing the mixing process leads to a noisy segmentation which severely affects accuracy. 

\begin{figure}[ht!]
    \centering        
    \setlength{\tabcolsep}{6pt}
    \begin{tabular}{cc}
        \includegraphics[width=.4\textwidth]{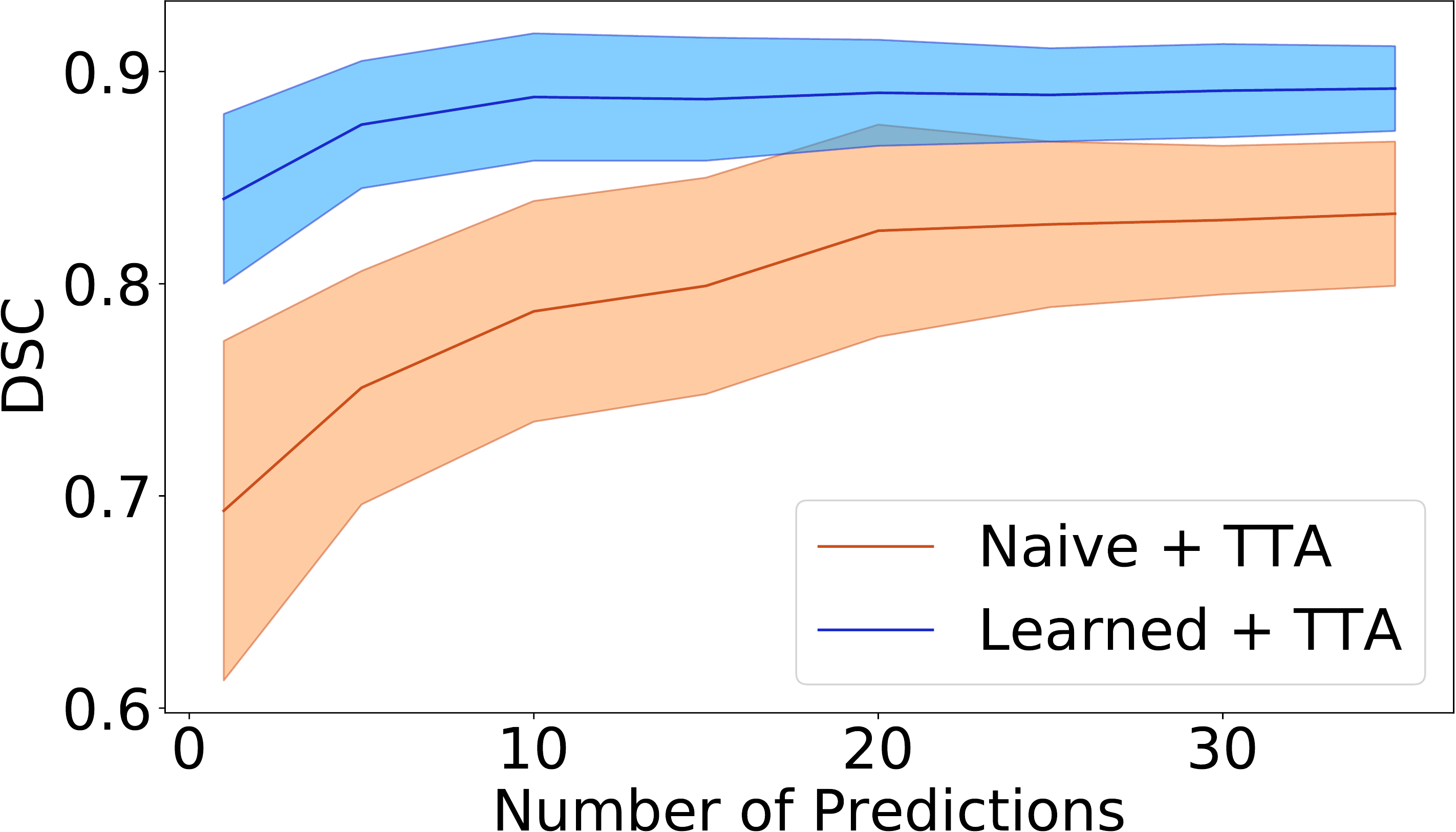}
        &\includegraphics[width=.4\textwidth]{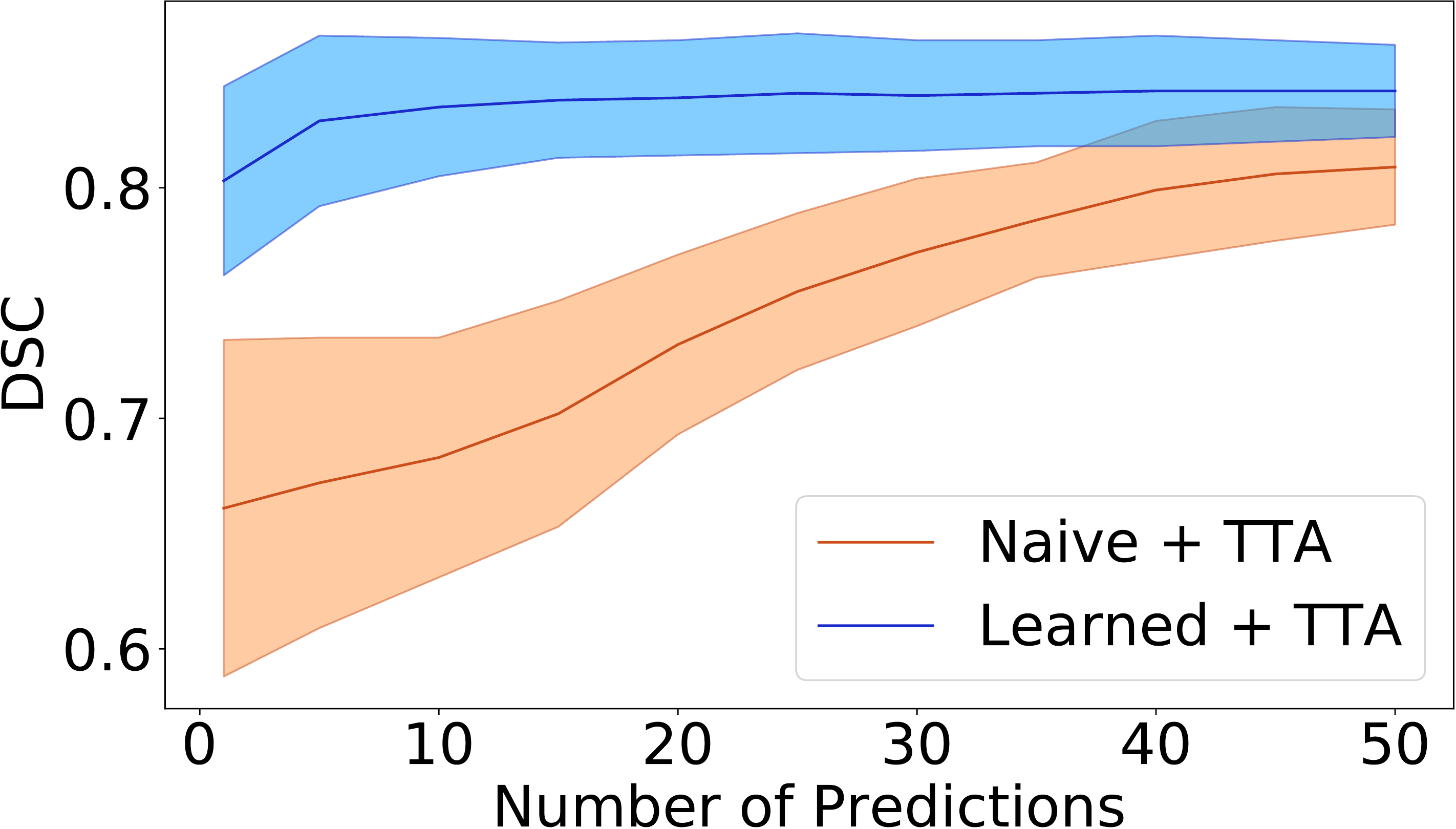}\\
        PPMI & BraTS2021
    \end{tabular}
    \vspace{-0.1cm}
    \caption{Segmentation accuracy (DSC) against the number of TTA predictions.}
    \vspace{-0.7cm}
    \label{fig:seg_acc_num_pred}
\end{figure}

Results in Table \ref{table:experiment_result} also demonstrate the positive impact of our TTA strategy on segmentation performance. Thus, adding this strategy to the naive unmixing approach increases the overall Dice by 13.2\% for PPMI and by 11.1\% for BraTS2021. Likewise, combining it with the learned unmixing approach boosts the overall Dice by 4.9\% for PPMI and by 3.7\% in the case of BraTS2021. Looking at the predictions for different reference patches in 
Fig.~\ref{fig:segmenation_patch}, we see a high variability, in particular for the naive unmixing approach. As can be seen in the first column of the figure (Naive\,+\,TTA and Learned\,+\,TTA), averaging multiple predictions in our TTA strategy reduces this variability and yields a final prediction very close to the ground-truth. As in other TTA-based approaches, our TTA strategy incurs additional computations since a segmentation prediction must be made for each augmented example (note that these predictions can be made in a single forward pass of the segmentation network). It is therefore important to analyze the gain in segmentation performance for different numbers of TTA augmentations. As shown in Fig.~\ref{fig:seg_acc_num_pred}, increasing the number of predictions for augmented examples leads to a higher Dice, both for the naive and learned unmixing approaches. Interestingly, when using the learned unmixing (i.e., Learned\,+\,TTA), the highest accuracy is reached with only 10-15 augmentations. In summary, our TTA strategy brings considerable improvements with limited computational overhead.

\mypar{Blind source separation} To assess whether our mixing-based image encoding effectively prevents an authorized person to recover the source image, we try to solve this BSS problem using the Deep Generative Priors algorithm introduced in \cite{DBLP:journals/corr/abs-2002-07942}. This algorithm uses a Noise Conditional Score Network (NCSN)~\cite{DBLP:journals/corr/abs-1907-05600} to compute the gradient of the log density function with respect to the image at a given noise level $\sigma$, ${\nabla}_x \log p_{{\sigma}}(x)$. An iterative process based on noise-annealed Langevin dynamics is then employed to sample from the posterior distribution of sources given a mixture. We use the U-Net++ as model for the NCSN, and train this model from scratch for each dataset with a Denoising Score Matching loss. Training is performed for 100,000 iterations on NVIDIA A6000 GPU, using the Adam optimizer with a learning rate of $5\times10^{\!-\!4}$ and a batch size of 16. 

\begin{figure}[t!]
        \centering
        \begin{footnotesize}
        \setlength{\tabcolsep}{3pt}
        \begin{tabular}{c|ccccc}
\includegraphics[width=0.12\textwidth]{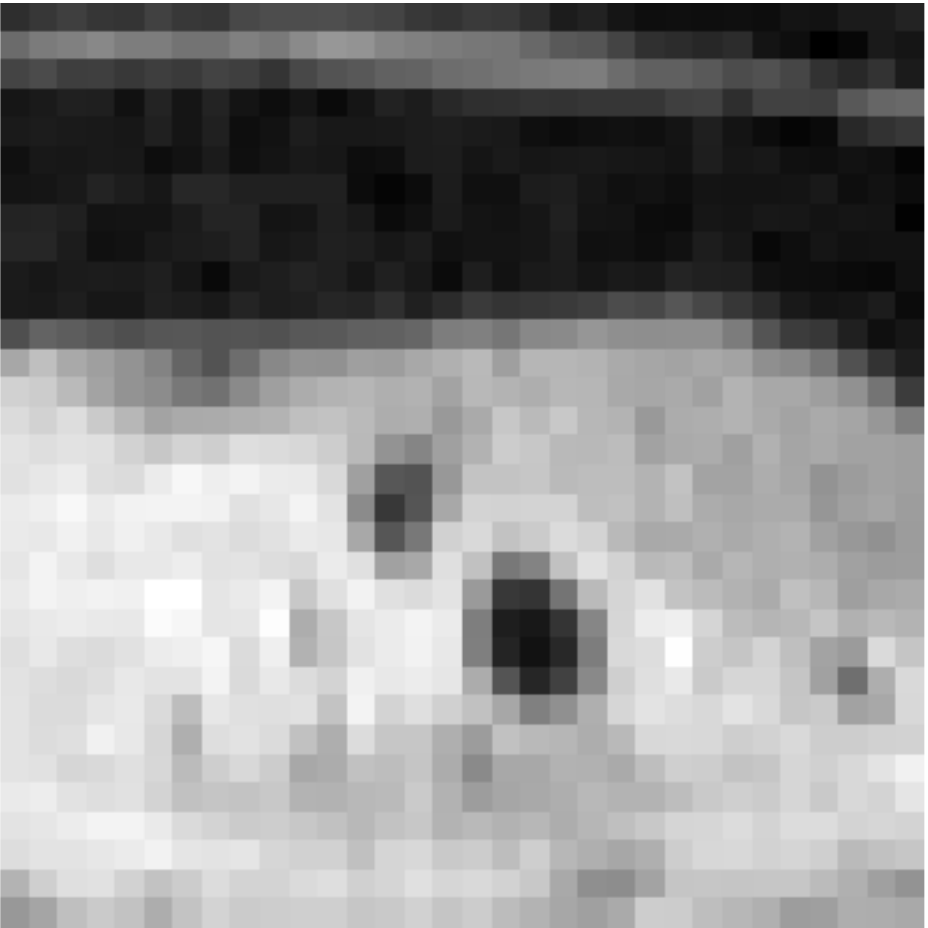}
& & \includegraphics[width=0.12\textwidth]{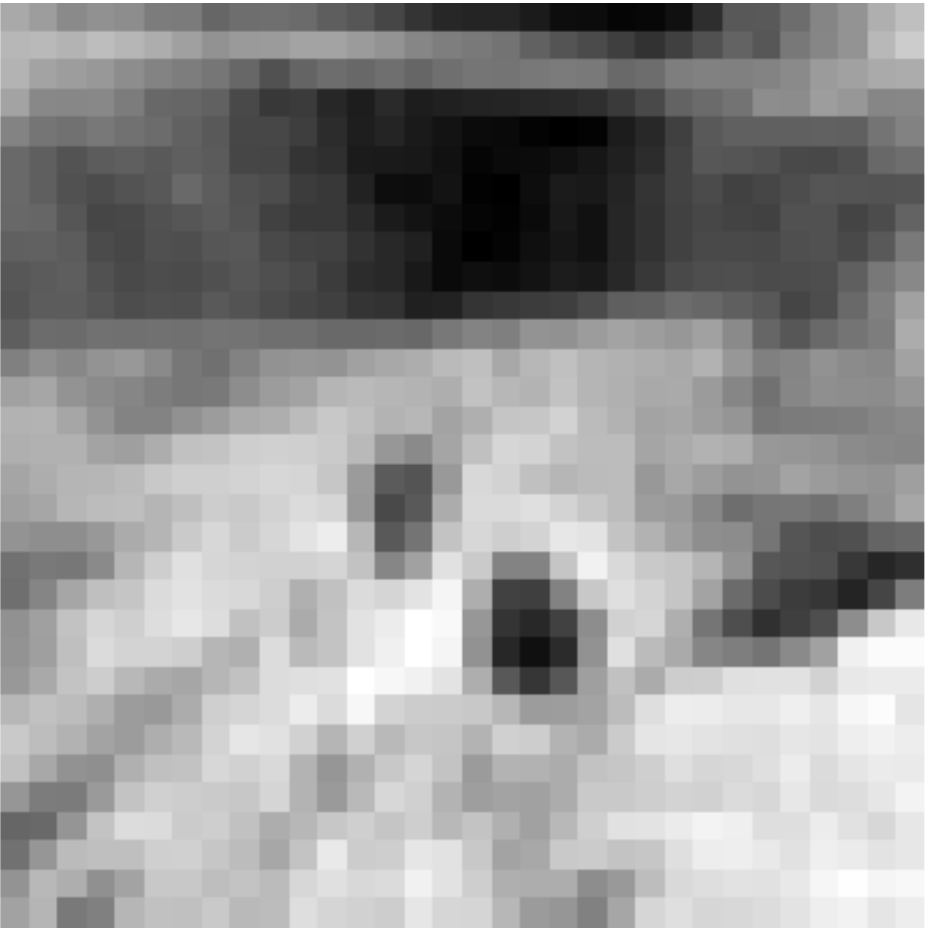}
&\includegraphics[width=0.12\textwidth]{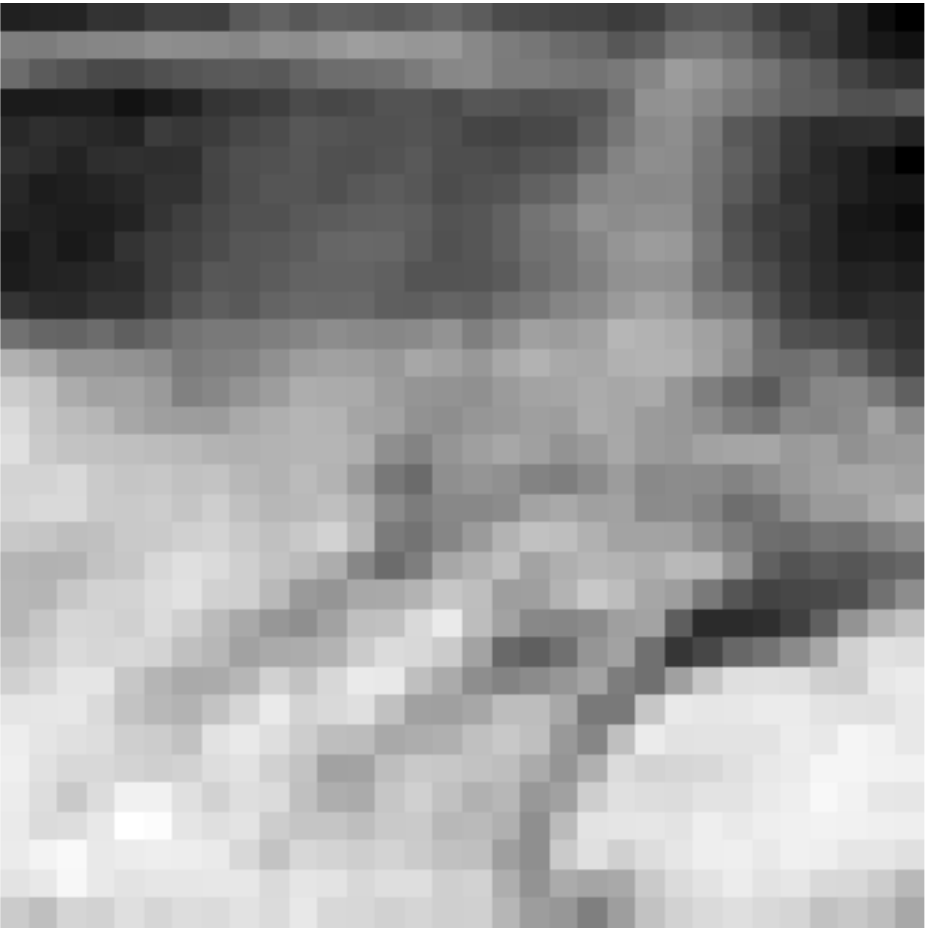}
&\includegraphics[width=0.12\textwidth]{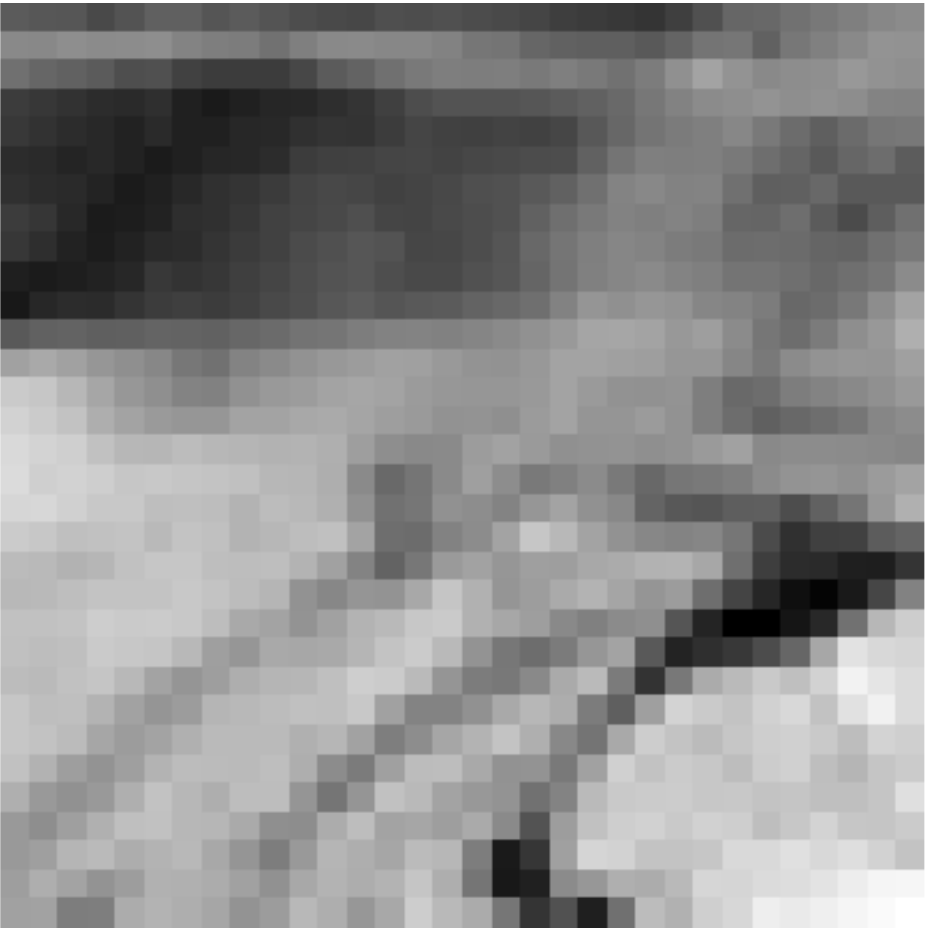}
&\includegraphics[width=0.12\textwidth]{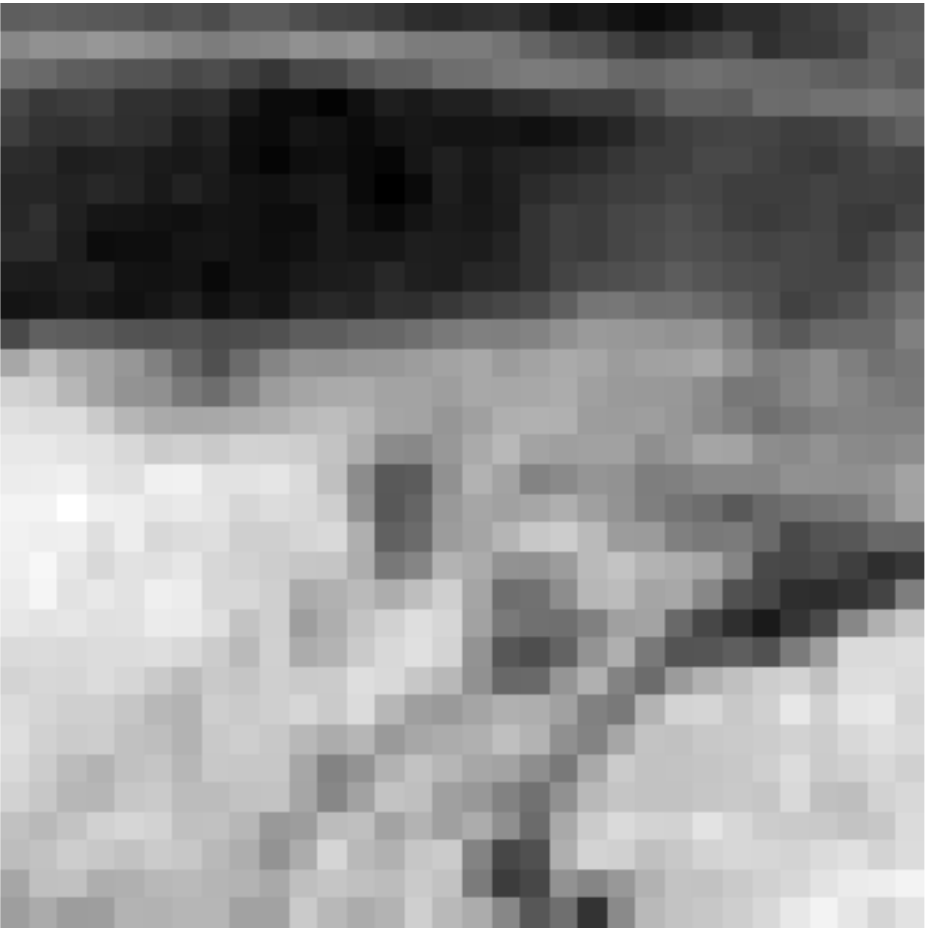}\\
Target (T) & & Pred. $\text{T}_1$ & Pred. $\text{T}_2$ & Pred. $\text{T}_3$ & Pred. $\text{T}_4$ \\[2.5pt]
\includegraphics[width=0.12\textwidth]{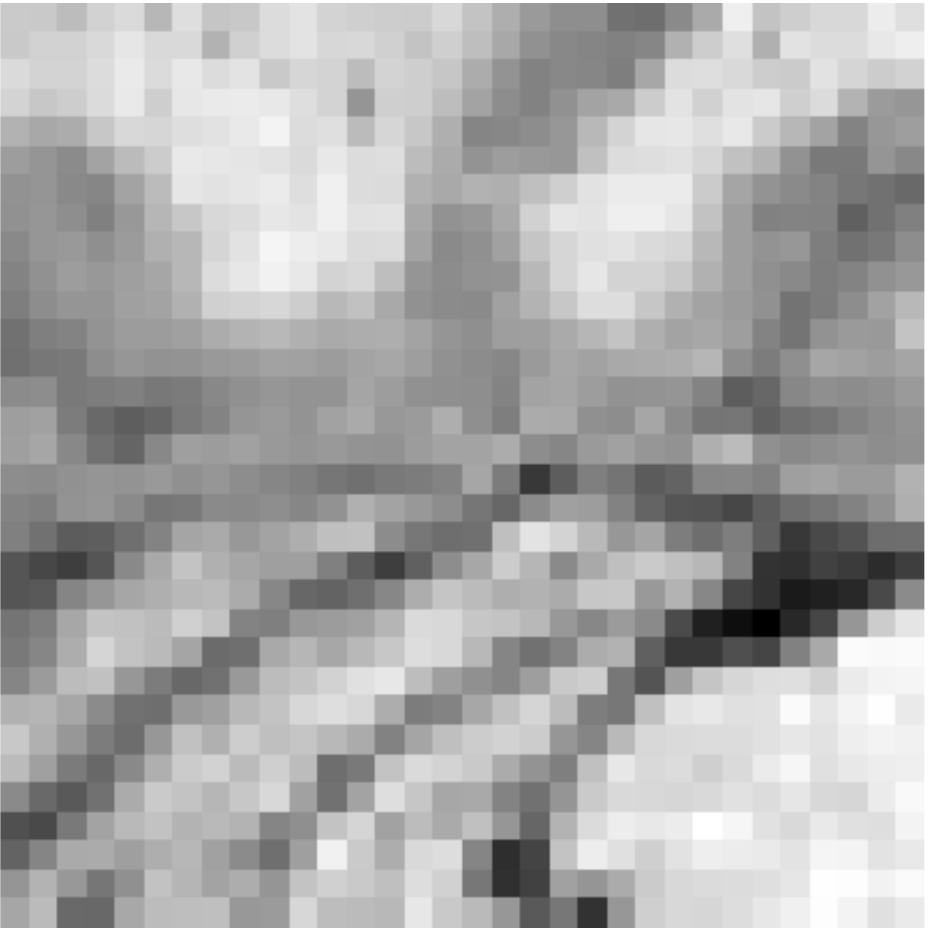}
& & \includegraphics[width=0.12\textwidth]{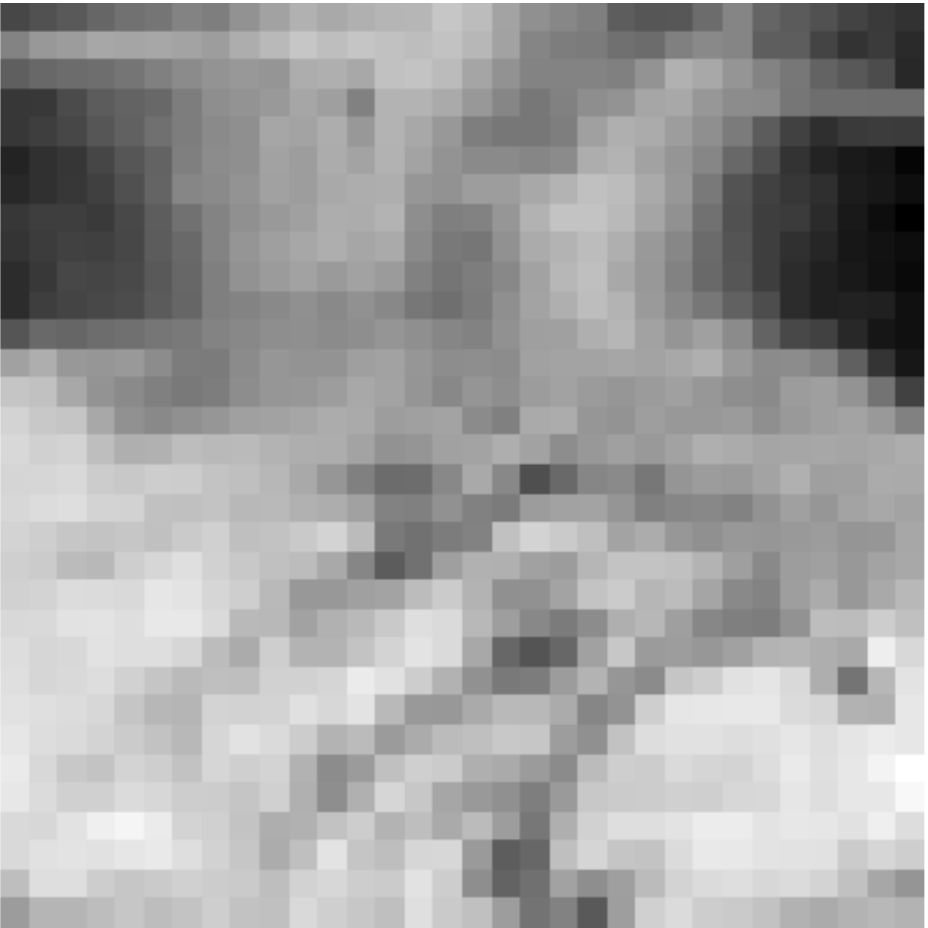}
&\includegraphics[width=0.12\textwidth]{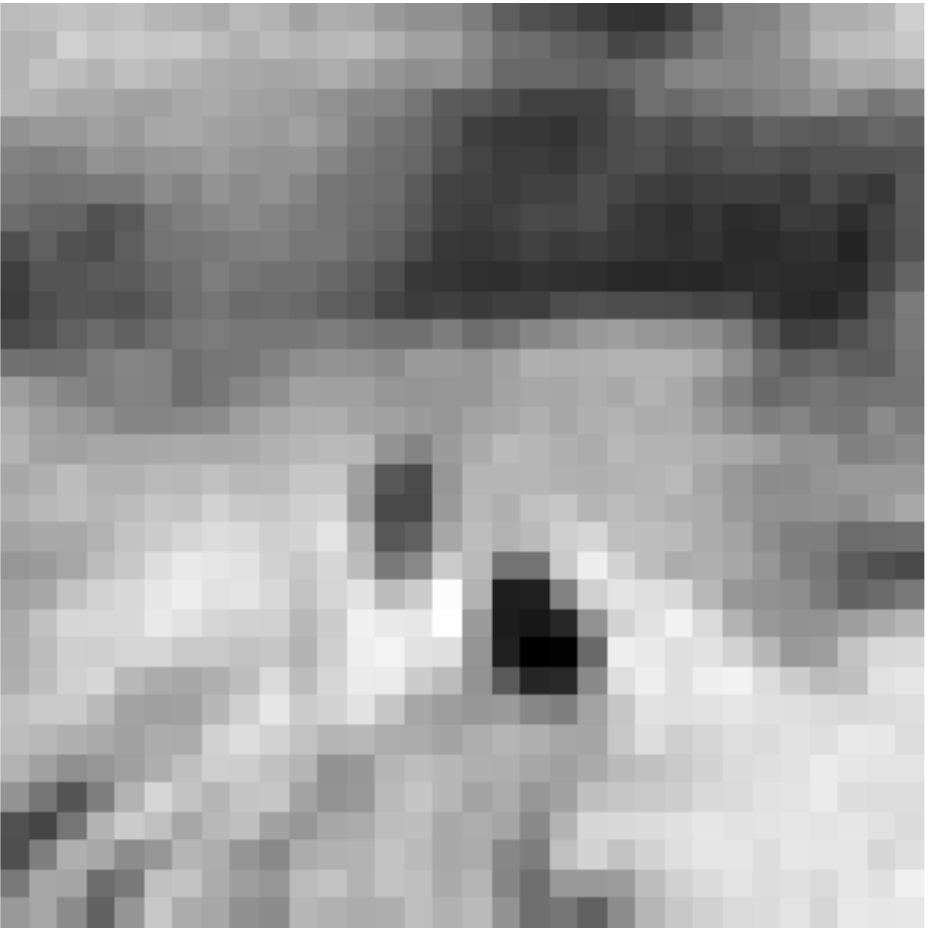}
&\includegraphics[width=0.12\textwidth]{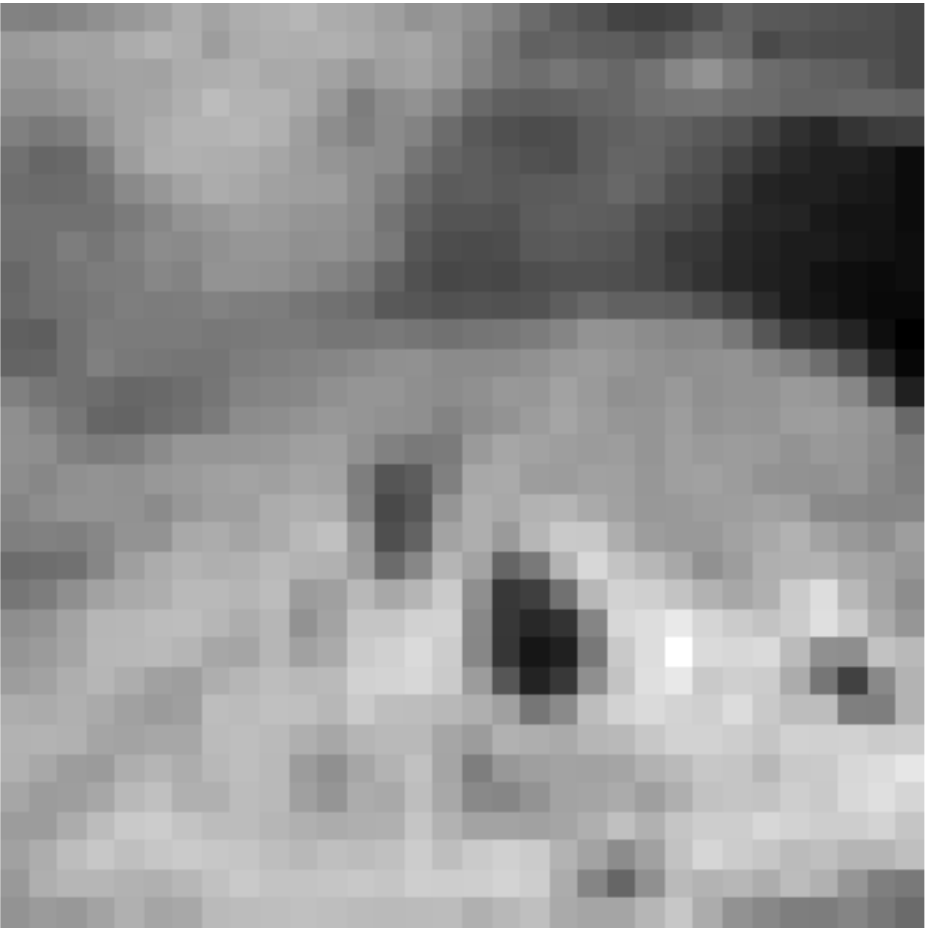}
&\includegraphics[width=0.12\textwidth]{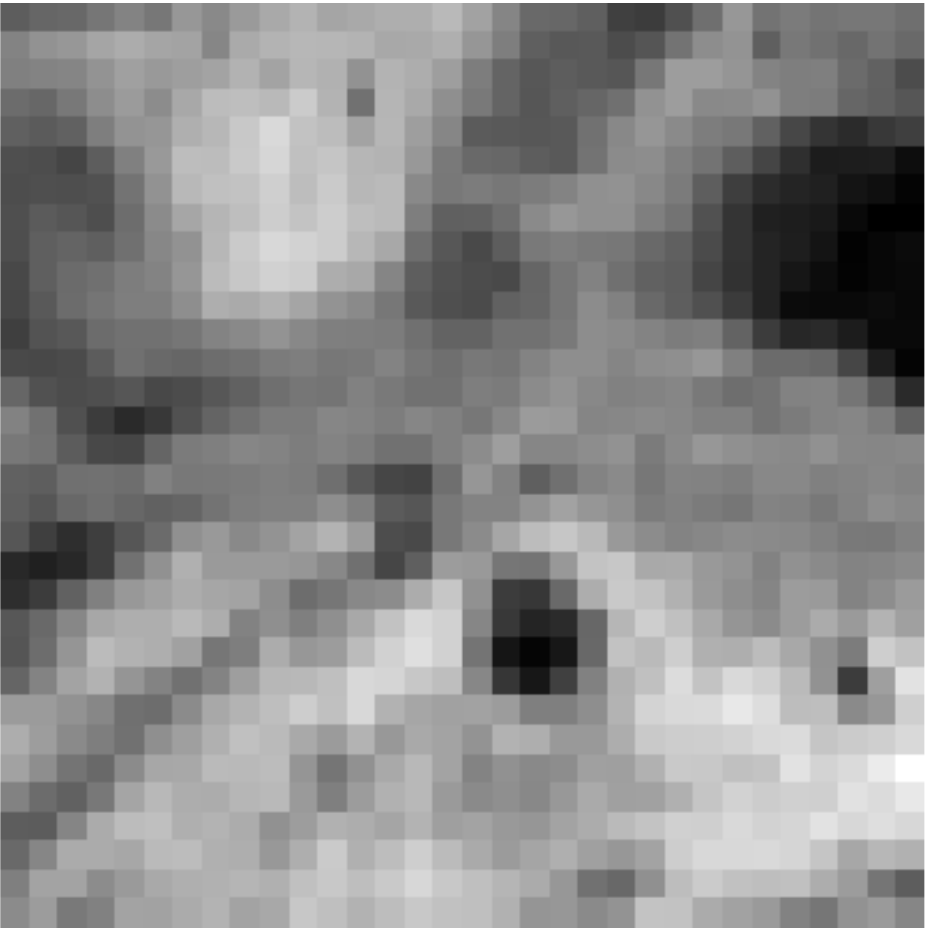}\\
Reference (R) & & Pred. $\text{R}_1$ & Pred. $\text{R}_2$ & Pred. $\text{R}_3$ & Pred. $\text{R}_4$ \\[2.5pt]
\includegraphics[width=0.12\textwidth]{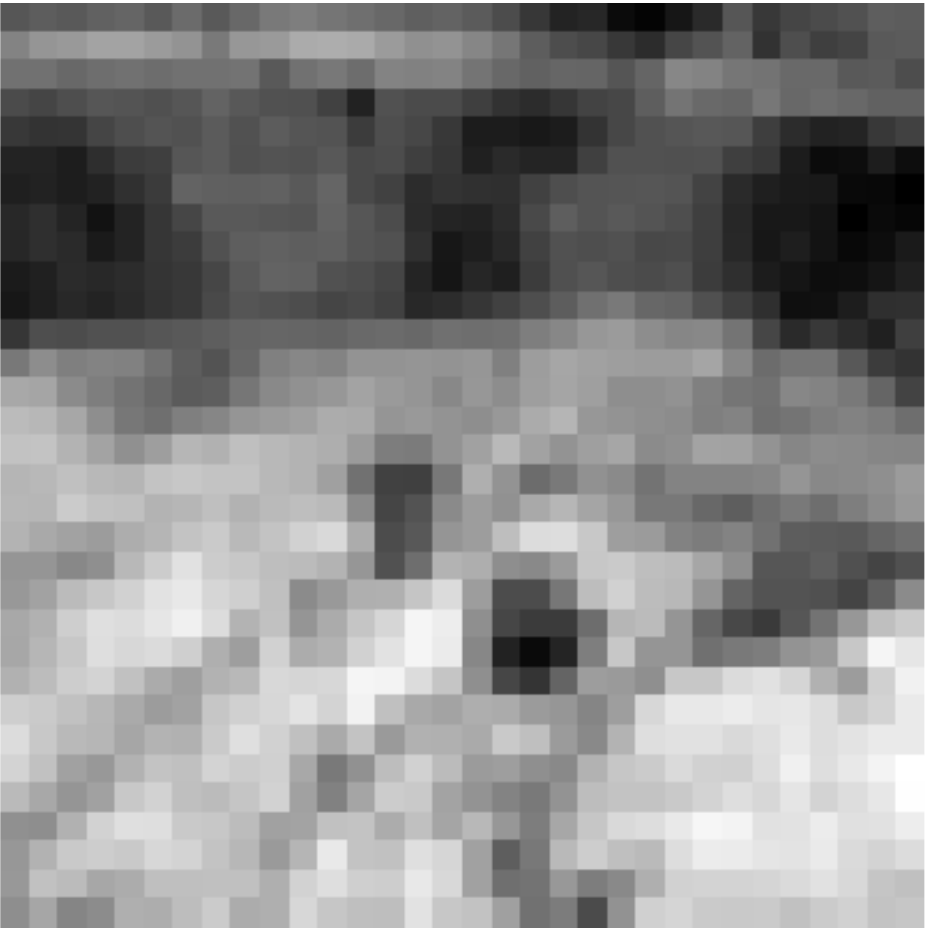}
& & \includegraphics[width=0.12\textwidth]{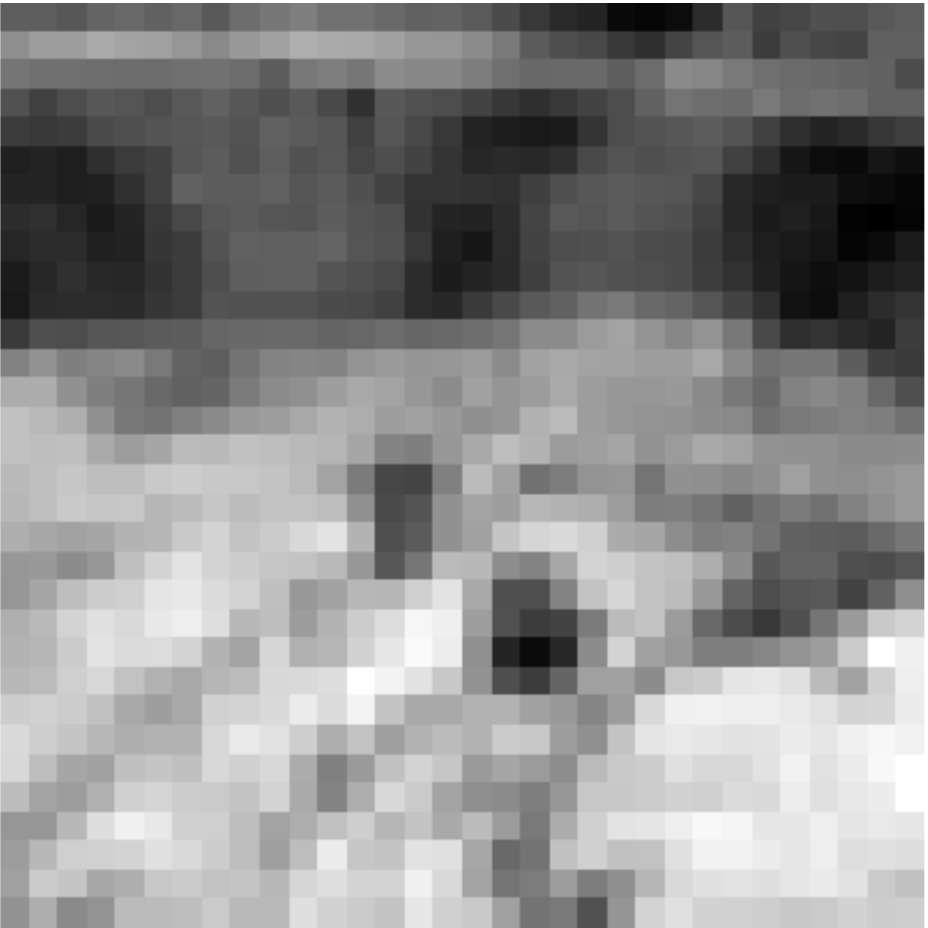}
&\includegraphics[width=0.12\textwidth]{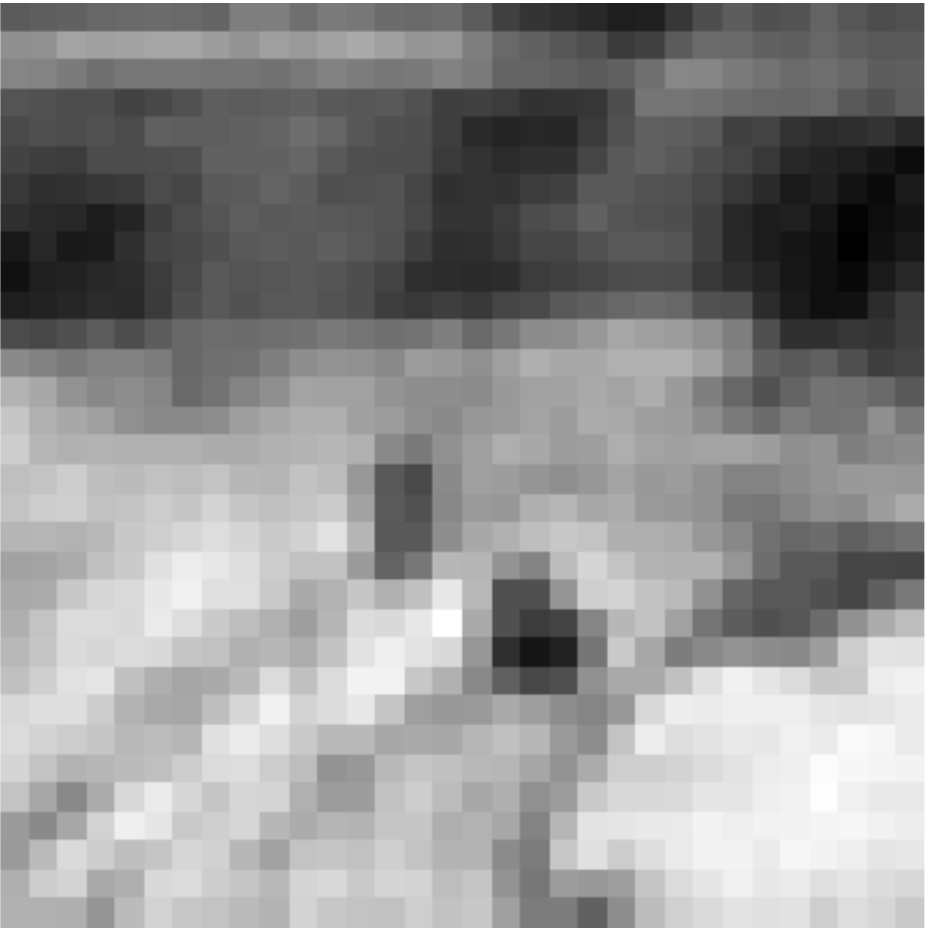}
&\includegraphics[width=0.12\textwidth]{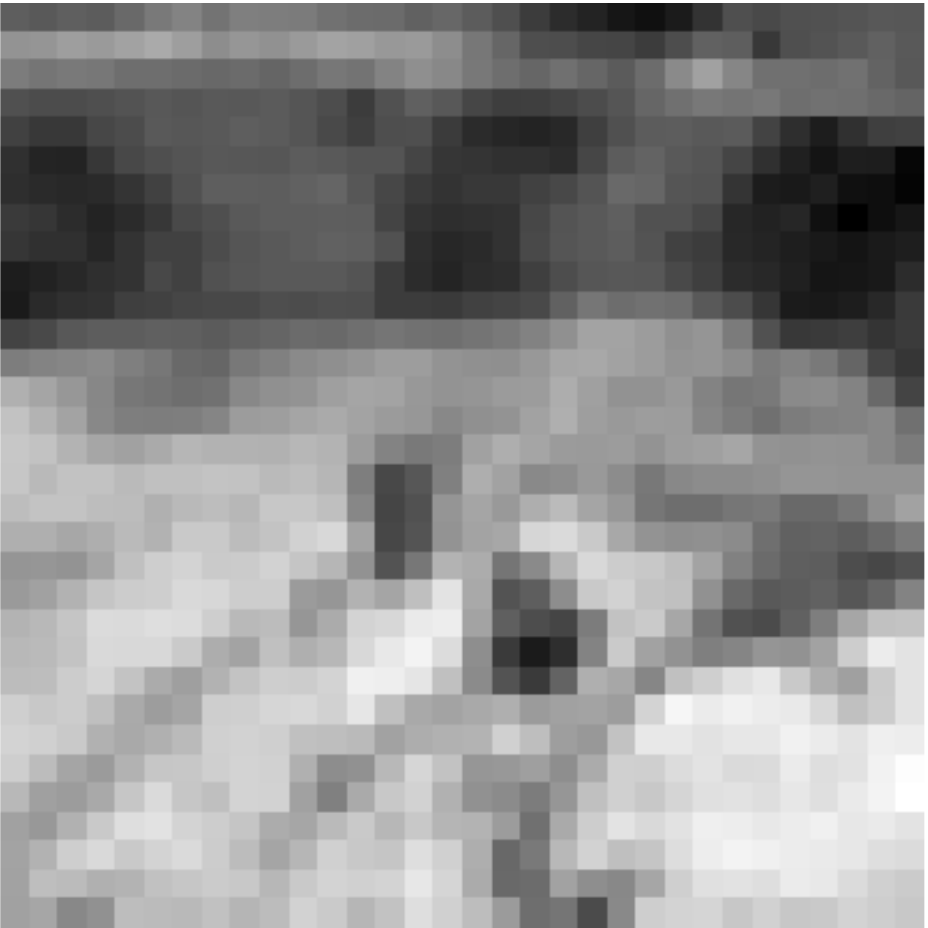}
&\includegraphics[width=0.12\textwidth]{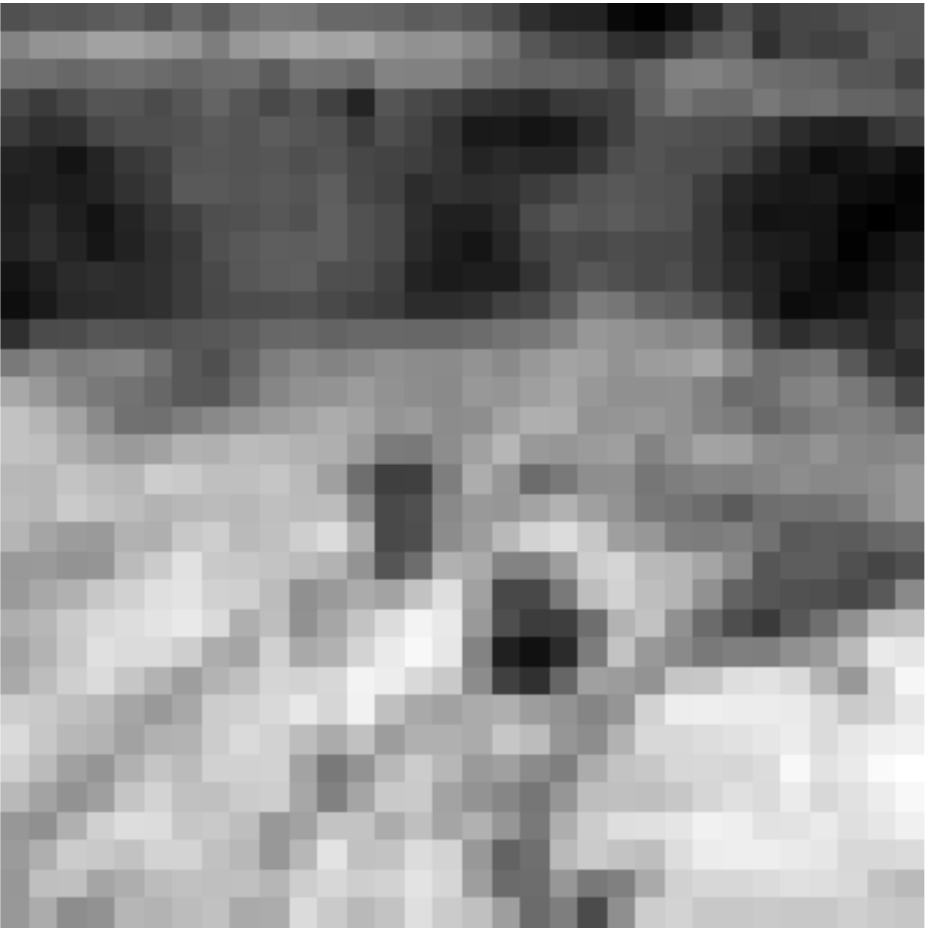}\\
Mixture (M) & & Pred. $\text{M}_1$ & Pred. $\text{M}_2$ & Pred. $\text{M}_3$ & Pred. $\text{M}_4$\\[-.5em]
        \end{tabular}
        \end{footnotesize}
        \caption{Examples of blind source separation (BSS) results for the mixture of given target and reference patches. Columns 2-5 correspond to results for different random initializations of the BSS algorithm.}
        \vspace{-0.4cm}
        \label{fig:patch_mix_unmix}
    \end{figure}

The second section of Table~\ref{table:experiment_result} gives the mean ($\pm$ stdev) of MS-SSIM scores (ranging from 0 to 1) between original target images and those recovered from the BSS algorithm: $0.602 \pm0.104$ for PPMI and $0.588\pm0.127$ for BraTS2021. These low values indicate that the target image cannot effectively be recovered from the mixed one. This is confirmed in Fig.~\ref{fig:patch_mix_unmix} which shows the poor separation results of the BSS algorithm for different random initializations.


\mypar{Test-retest reliability} One source of variability in our method (without TTA) is the choice of the reference image used for mixing. To evaluate the stability of our method with respect to this factor, we perform a test-retest reliability analysis measuring the intra-class correlation coefficient (ICC) \cite{article} of the test DSC for two predictions using different references. A higher ICC (ranging from 0 to 1) corresponds to a greater level of consistency. The third section of Table~\ref{table:experiment_result} reports the ICC score obtained for each segmentation class, as well as the upper and lower bounds at 95\% confidence. We see that all ICC values are above 0.75, indicating a good reliability.

\setlength{\intextsep}{7pt}%
    \begin{wraptable}{r}{6cm}
    \centering
    \caption{Subject re-identification analysis on the PPMI dataset.\vspace{3pt}}
    \label{table:Subject_identification}
    \small
    \setlength{\tabcolsep}{4pt}
        \begin{tabular}{lcc}
        \toprule
        Method & F1-score & mAP\\
        \midrule
        No Proxy &  0.988 & 0.998 \\
        Privacy-Net~\cite{kim2020privacynet} &  0.092 & 0.202 \\
        Deformation-Proxy~\cite{bach2021nonlinear} & 0.122 & 0.147 \\
        Ours & 0.284 & 0.352 \\
        \bottomrule
        \end{tabular}
    \end{wraptable}

    
\mypar{Subject re-identification} To measure how well our method protects the identity of patients, we carry out a patient re-identification analysis using the PPMI dataset which has multiple scans for the same patient. In this analysis, we encode each image in the dataset by mixing it with a randomly chosen reference. For an encoded image $x_{\mr{mix}}$, we predict the patient identity as the identity of the other encoded image $x'_{\mr{mix}}$ most similar to $x_{\mr{mix}}$ based on the MS-SSIM score. Table~\ref{table:Subject_identification} compares the F1-score and mAP performance of our method to a baseline with no image encoding (No Proxy), Privacy-Net and Deformation-Proxy. As can be seen, the re-identification of patients is quite easy when no encoding is used (mAP of $0.998$), and all encoding-based methods significantly reduce the ability to recover patient identity using such retrieval approach. While our mixing based method does not perform as well as the more complex Privacy-Net and Deformation-Proxy approaches, it still offers a considerable protection while largely improving segmentation accuracy (see Table \ref{table:experiment_result}). 
\vspace{-0.2cm}
\section{Conclusion}
\vspace{-0.2cm}
We introduced an efficient method for privacy-preserving segmentation of medical images, which encodes 3D patches of a target image by mixing them to reference patches with known ground-truth. Two approaches were investigated for recovering the target segmentation maps from the mixed output of the segmentation network: a naive approach reversing the mixing process directly, or using a learned unmixing model. We also proposed a novel test-time augmentation (TTA) strategy to improve performance, where the image to segment is mixed by different references and the predictions for these mixed augmentations are averaged to generate the final prediction. 
 
We validated our method on the segmentation of brain MRI from the PPMI and BraTS2021 datasets. Results showed that using a learned unmixing instead of the naive approach improves DSC accuracy by more than 14\% for both datasets. Our TTA strategy, which alleviates the problem of prediction variability, can also boost DSC performance by 3.7\%-13.2\% when added on top of its single-prediction counterpart. Compared to state-of-art approaches such as Privacy-Net and Deformation-Proxy, our method combining learned unmixing and TTA achieves a significantly better segmentation, while also offering a good level of privacy.


In the future, we plan to validate our method on other segmentation tasks involving different imaging modalities. While we encoded a target image by mixing it to a reference one, other strategies could be also explored, for example, mixing more than two images. This could make the BSS more difficult, hence increasing the security of the method, at the cost of a reduced segmentation accuracy. The prediction variance of our TTA strategy could also be used as a measure of uncertainty in semi-supervised segmentation settings or to suggest annotations in an active learning system.   





%
%
%
\vspace{-0.2cm}

\footnotesize
\bibliographystyle{splncs04}
\bibliography{mybibliography}
%




\end{document}